\documentclass[10pt,twocolumn,letterpaper]{article}

\usepackage{wacv}              % To produce the CAMERA-READY version
\usepackage{graphicx}
\usepackage{amsmath}
\usepackage{amssymb}
\usepackage{booktabs}

\usepackage[accsupp]{axessibility}  % Improves PDF readability for those with disabilities.

\usepackage{tikz}
\usepackage{comment}
\usepackage{color}
\usepackage{multirow}
\usepackage{array}
\usepackage{graphbox}
\usepackage{url}
\usepackage{algorithm}
\usepackage{algpseudocode}
\usepackage{booktabs}
\newcommand*{\ShowNotes}{}
\definecolor{darkred}{rgb}{0.7,0.1,0.1}
\definecolor{darkgreen}{rgb}{0.1,0.7,0.1}
\definecolor{cyan}{rgb}{0.7,0.0,0.7}
\definecolor{dblue}{rgb}{0.2,0.2,0.8}
\definecolor{maroon}{rgb}{0.76,.13,.28}
\definecolor{burntorange}{rgb}{0.81,.33,0}
\ifdefined\ShowNotes
  \newcommand{\colornote}[3]{{\color{#1}\bf{#2: #3}\normalfont}}
\else
  \newcommand{\colornote}[3]{}
\fi

\definecolor{redmethod}{RGB}{192,0,0}
\definecolor{greenmethod}{RGB}{146,208,80}
\definecolor{bluedashed}{RGB}{68,114,196}
\definecolor{orangedashed}{RGB}{237,125,49}
\definecolor{greenscatter}{rgb}{0.0,0.5,0.0}
\definecolor{purplescatter}{rgb}{0.5,0.0,0.5}
\definecolor{orangescatter}{rgb}{1.0,0.67,0.0}
\definecolor{grayscatter}{rgb}{0.5,0.5,0.5}

\usepackage[pagebackref,breaklinks,colorlinks]{hyperref}

\usepackage[capitalize]{cleveref}
\crefname{section}{Sec.}{Secs.}
\Crefname{section}{Section}{Sections}
\Crefname{table}{Table}{Tables}
\crefname{table}{Tab.}{Tabs.}

\def\wacvPaperID{238} % *** Enter the WACV Paper ID here
\def\confName{WACV}
\def\confYear{2024}

\begin{document}

%%%%%%%%% TITLE
\title{CLID: Controlled-Length Image Descriptions with Limited Data}

\author{Elad Hirsch \hspace{0.3in} Ayellet Tal\\
Technion -- Israel Institute of Technology\\
{\tt\small \{eladhirsch@campus,ayellet@ee\}.technion.ac.il}
}

\maketitle

% \thispagestyle{empty}

%%%%%%%%% TEASER
% figure
% \teaser{
\begin{figure*}[htb]
    \begin{minipage}[b]{0.16\linewidth}
    \raisebox{-0.4\height}{\hspace{-0.1cm}\includegraphics[width=1.4\linewidth]{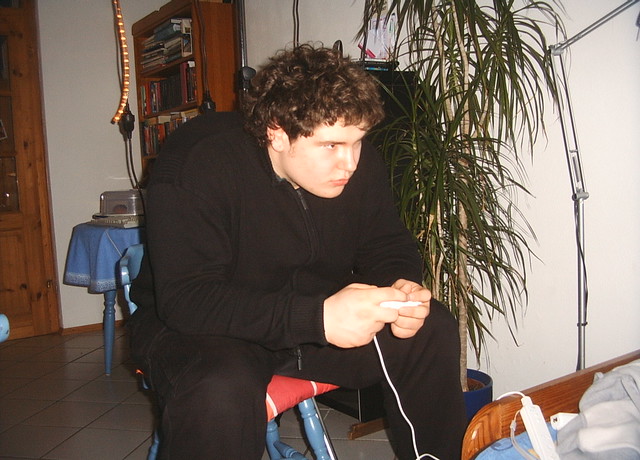}}
    \end{minipage}
    \begin{minipage}[b]{0.83\linewidth}
    %\centering
    \begin{flushright}
    \resizebox{0.96\textwidth}{!}{%
    \hspace{0.5cm}
    \begin{tabular}{m{0.09\linewidth} m{0.11\linewidth}  m{0.85\linewidth}}
     \toprule
     % \multirow{3}{*}{\raisebox{-0.95cm}{\cite{deng2020length}}} & Level 1 &  a young man plays with a videorest. \\
     \multirow{3}{*}{\raisebox{-1.2cm}{\shortstack[l]{LaBERT \\[0.02in] \cite{deng2020length}}}} & Level 1 &  a young man plays with a videorest. \\
     \cmidrule{2-3}
     & Level 3 & a young man sitting in a living room with a wii remote and a white rug on his head. \\
     \cmidrule{2-3}
     & Level 6 & a boy sitting in a chair next to a table with a towel on it and a surfboard on the floor next to a book shelf with a book case on the table and a surfboard on top of the table and a plant on the wall. \\
     \midrule[0.8pt]
     \multirow{3}{*}{\raisebox{-1.2cm}{\shortstack[l]{CLID \\ (ours)}}} & Level 1 & a young boy is playing a video game. \\
     \cmidrule{2-3}
     & Level 3 & a boy is playing a video game in a living room with a plantted plant in the background. \\
     \cmidrule{2-3}
     & Level 6 & a boy in a black jacket with dark brown hair is holding a wii remote in a room area with a stack of shelves and a chair with a blue table and a bookcase next to a bed and a cord on to the side of the bed. \\
     \midrule[0.8pt]
     & (a) Length & (b) Generated Caption
    \end{tabular}}
    \end{flushright}
    \end{minipage}
\vspace{-0.08in}
\caption{\textbf{Length-controlled image captioning.}
People describe a given image briefly or in length.
Most previous works generate short captions, which are prevalent in existing datasets.
We propose a method that generates captions of sought-after lengths.
Our method generates long captions, which hardly exist in training datasets, and  achieves comparable results to SoTA methods for short captions.
% We can even generate captions of lengths that do not exist in the dataset, which previous works could not.
}
\label{fig:teaser}
\vspace{-0.08in}
\end{figure*}

\begin{abstract}
Controllable image captioning models generate human-like image descriptions, enabling some kind of control over the generated captions.
This paper focuses on controlling the caption length, i.e. a short and concise description or a long and detailed one.
Since existing image captioning datasets contain mostly short captions, generating long captions is challenging.
To address the shortage of long training examples, we propose to enrich the dataset with varying-length self-generated captions.
These, however, might be of varying quality and are thus unsuitable for conventional training.
We introduce a novel training strategy that selects the data points to be used at different times during the training.
Our method dramatically improves the length-control abilities, while exhibiting SoTA performance in terms of caption quality.
Our approach is general and is shown to be applicable also to paragraph generation. 
Our code is publicly available~\footnote{https://github.com/eladhi/CLID}.
\end{abstract}

%%%%%%%%%%%%%%%%%%%%%%%%%%%
%% section: Introduction
%%%%%%%%%%%%%%%%%%%%%%%%%%%
\section{Introduction}
\label{sec:introduction}

{\em Image Captioning} refers to the task of generating human-like image descriptions~\cite{vinyals2015show}.
It relies on supervision, utilizing large datasets of image and text pairs~\cite{chen2015microsoft,hodosh2013framing,young2014image}.
% Traditionally, captioning models are trained on the training split and evaluated on the test split, with respect to several metrics.
%
\textit{Controllable Image Captioning (CIC)} aims at generating a caption while satisfying a constraint or a user request.
The constraint may 
% have the form of an input control signal, that can 
relate to the content~\cite{cornia2019show,zhong2020comprehensive}, the style~\cite{mathews2018semstyle}, the structure~\cite{chen2021human}, or the length~\cite{deng2020length}  of the caption.

Our work focuses on controlling the description length and consequently, the amount of detail in it.
% Our goal is to eliminate this length limitation and allow caption-length flexibility.
This may suit different people/applications at different scenarios.
For instance, visually-impaired observers may prefer a short and concise description when in a hurry and a long and detailed one at other times.
Other applications, such as text-based image retrieval, may also benefit from this flexibility.
% The paper focuses on image captioning, nevertheless we demonstrate the generality of our method by presenting results for  paragraph generation, which is a different yet related task.

Captioning training datasets have length limitations, as human-annotated captions tend to be short, concise and omit some visual information
%, since  annotators often focus on the essence of the image
% although  they are rich in objects' relations objects
~\cite{farhadi2010every}.
For example, in MS-COCO~\cite{chen2015microsoft} $95\%$ of the training captions contain less than $15$ words.
This limits the length of the generated captions, making the generation of long captions a challenge.
% Descriptive paragraphs, on the other hand, suffer from the opposite problem---the descriptions are long and generating short paragraphs is challenging~\cite{krause2016paragraphs}.
%
% SoTA models require a lot of data to learn from.
% % This, unfortunately is not the case.
The straightforward solution of collecting image descriptions of varying lengths is infeasible.
% (For paragraphs,  enriching the dataset by subsets of sentences  will not yield a satisfactory dataset, since the order of the sentences matters.)
%But, that operation is not feasible for many researchers and institutions.
This paper shows how to
% to utilize SoTA models  to control the description length, addressing that shortage of data.
address this shortage of data.

We propose a novel approach for generating varying-length image descriptions with inadequate training data.
% Our solution is general and may be applied to both tasks.
% For simplicity, throughout the paper, we discuss the more common task of captioning.
% In Section~\ref{sec:results} (results), we show that our approach is as beneficial for paragraph generation.
%
Our approach is based on solving two sub-problems.
At first, we automatically generate long "synthetic" captions, termed {\em self-generated} captions.
While this procedure addresses the lack of long training captions, the generated captions might be of low quality ({\em domain gap}).
This leads to the second challenge of how to use this low-quality dataset during training.
We present a novel training method that makes use of the trusted dataset of short captions jointly with the large varying-quality  caption dataset, in an informed manner.

To address the first challenge, we propose a method that parses scene graphs and generates varying-length sentences.
Scene graphs contain the essential descriptive components: objects, objects' attributes, relations, and activities~\cite{farhadi2010every,johnson2015image}.
Image saliency~\cite{JIA20EML} is utilized for generating reasonable sentences.
We end up with a large dataset of self-generated varying-length \& varying-quality descriptions.

The second challenge is how to use the {\em extended  dataset}, which is composed of the trusted short captions and the self-generated captions,  during training.
Since this dataset might contain inaccuracies (that exist in the scene graphs), repetitions, and different linguistic styles, we cannot consider both caption types similarly.
% difference from that of the short ground-truth captions.
%
We introduce a training procedure in which  the use of the self-generated data is guided  by the
(small and short-caption) 
trusted dataset.
In particular, our method progressively filters out low-quality captions  during training.
It assures that even though the very long captions might be eliminated early on and the remaining high-quality captions are mostly short, the model remembers enough of the long captions, to be able to generate ones.
For this strategy to work, we should be able to measure caption quality, a topic that is addressed as well.

We demonstrate the benefits of our approach on the MS-COCO Caption dataset~\cite{chen2015microsoft}, in which
% , which contains $123,287$ images and $5$ human-written captions per image.
%The length distribution in this dataset is biased towards short captions, i.e. $11.8\%$ of the training captions are of size 1-9 tokens, $86.6\%$ are of size 10-19 and the remaining $1.6\%$ are of size 20-55 tokens.
the mean caption length is $10.47$ words. %, with standard deviation of $2.4$.
% In contrast, in our extended dataset almost $40\%$ of the captions are larger than 20 tokens.
% The mean length rises to $21.3$ with a standard deviation of $13.56$.
For long captions, which barely appear in the dataset, our work highly improves the control precision (by up to $27\%$), while maintaining the caption quality, as demonstrated in Figure~\ref{fig:teaser}.
It even 
% Furthermore, our method is the only one that 
enables generating captions of lengths that do not appear in the dataset at all.
%
% For a given image our model generates captions of different lengths (levels $1$-$7$).
The short captions contain only the essence of the image, whereas the long captions reveal many more details.
% For descriptive paragraph generation, we demonstrate similar benefits of our approach on the dataset of \cite{krause2016paragraphs}.
% We name our method {\em LCSD} (\textbf{L}ong image \textbf{C}aptions from \textbf{S}hort caption \textbf{D}atasets).

Our method is general and may be applied also to the task of {\em descriptive paragraph generation} from images~\cite{krause2016paragraphs}, which is a related, yet distinct task.
While both tasks generate a coherent natural-language image description, the linguistic structure and the amount of details differ.
Captioning refers to a single descriptive sentence, whereas paragraphs consist of multiple sequential sentences.
% The tasks are evaluated on different datasets.
The existing paragraph dataset~\cite{krause2016paragraphs} is small and contains mostly long descriptions.
In Section~\ref{sec:results} we show that despite the very small dataset, our method  manages to generate length-control results for this task as well.

Hence, our contributions are as follows:
\vspace{-0.05in}
\begin{enumerate}
    % \vspace{-0.12in}
    \item 
    We introduce a novel approach for length-controllable image captioning.
    It manages to generate long, even out-of-distribution, descriptions. % despite having a dataset that contains mostly short captions.
%
%    For short captions, which were addressed in previous works, it achieves similar control precision and quality.
    \vspace{-0.105in}
    \item    
     Since our approach is general and unified, it can  be used for paragraph generation.
     This is the first work to introduce length-control abilities for this task.
    \vspace{-0.105in}
    \item
    We present a method for generating varying-length diverse-quality captions from scene graphs, without the need of a ground-truth dataset.
    \vspace{-0.105in}
    \item
    We propose a training procedure that learns both from high-quality data and from low-quality data, such that the essence of the low-quality data is not forgotten.
\end{enumerate}

%%%%
% Topics to cover:
% Image captioning
% Controllable image captioning, specifically length
% SGs for captioning (just say it's different - they still use COCO)
% Creating a noisy dataset, with grounding
% denoising a dataset
%%%%%%%%%%%%%%%%%%%%%%%%%%%
%% section: Related Work
%%%%%%%%%%%%%%%%%%%%%%%%%%%
\section{Related work}
\label{sec:RelatedWork}
%We provide a brief background on the relevant domains.

%\vspace{0.03in}
\noindent
{\bf Image captioning.}
Image captioning is a core task in scene understanding.
The common approach in recent years combines a visual encoding model, which extracts visual features, with a language model that learns to generate text from both the visual features and the input text.
The visual encoding model extracts global image features~\cite{donahue2015long,vinyals2015show}, patch features 
%(as a grid of CNN features)
~\cite{wang2017skeleton,xu2015show} or region features~\cite{anderson2018bottom,qin2019look}.
Other image representation structures  contain the visual features jointly with additional relevant information, such as scene graphs~\cite{nguyen2021defense,yang2019auto}.
%, which contain complementary object names and relationships between objects.
The language models also vary, where
RNNs~\cite{lu2018neural,vinyals2015show} and transformers~\cite{cornia2020meshed,Herdade2019transforming,zhang2021vinvl} are common choices.
%For more information, please refer to an excellent review of the field~\cite{stefanini2021show}.

%\vspace{0.03in}
\noindent
{\bf Controllable image captioning (CIC).}
This task takes captioning to the next level by adding constraints, which are usually user-defined.
% The goal is to achieve high accuracy while satisfying the requirements.
Such constraints may relate to caption linguistic style~\cite{alikhani2020clue,chen2018factual},  content~\cite{chen2020say,cornia2019show,kim2019dense,zhong2020comprehensive} or  structure~\cite{chen2021human,deshpande2019fast}.
Our work focuses on length control, also addressed in~\cite{deng2020length} and partially in~\cite{kastner2021imageability}.
% The work of~\cite{deng2020length} (LaBERT) was the first to tackle the length controllable captioning.
In~\cite{deng2020length} an interesting architecture, {\em LaBERT}, is proposed for the problem.
This is a transformer-based captioning model.
%, as illustrated in Fig.~\ref{fig:labert}.
Its input consists of visual and word embeddings, where
%the visual embeddings represent image regions and their position in the image and 
%the word embeddings represent the words, their position and the caption length.
the word embedding also represents the caption length.
The length is measured in {\em tokens} (sub-words), 
%e.g, playing = play + ing
which appear in the BERT vocabulary~\cite{devlin-etal-2019-bert}.
%rather than natural words. 
% The desired length is given in {\em length levels}, which correspond to lower and upper length bounds.
% For instance, level $1$ may correspond to [$1$-$9$] tokens and level $2$ to [$10$-$19$].
This model manages to control the length of captions up to $30$ tokens well, with a precision rate of above $90\%$.
Such captions are $99.8\%$ of the training dataset.
For longer lengths, which are the remaining $0.2\%$, the precision drops by 10-43\%.
%However, due to the caption length distribution in the dataset, they focus on captions of up to $25$ tokens long.
% We use their architecture for controlling length, but propose a novel approach for self-generating caption for training, in order to support captions of up to 69 tokens, even if such captions are limited in the original training dataset or does not exist in it at all.
The work of~\cite{kastner2021imageability} aims to control both the image caption length and imageability (i.e., the clarity of the mental image).
In the context of length, they use~\cite{deng2020length}.
% , applied on the original dataset with all its limitations.
% For imageability, they generate imageability embeddings to be added as input to LaBERT.
% These works therefore do not aim to tackle the lack of long captions, which is our focus.
Our method addresses the limitation of long caption generation.

%\vspace{0.03in}
\noindent
{\bf Descriptive image paragraphs.}
Paragraph generation aims at describing an image by a sequence of sentences.
Common works use hierarchical RNNs to guide the sentence topics (high level) and word sequence (low level)~\cite{chatterjee2018diverse,krause2016paragraphs,mao2018show,wang2019convolutional,zha2019context}.
Other solutions further guide the training of RNN models with reinforcement learning~\cite{luo2019curiosity}, scene graph hierarchy information~\cite{yang2020hierarchical} and adversarial training~\cite{liang2017recurrent}.
Despite the resemblance to single-sentence captioning, the tasks are considered distinct and thus the training datasets differ.
Length-controllable paragraph generation has not yet been addressed.

%\vspace{0.03in}
\noindent
{\bf Scene graphs.}
A scene graph represents the content of an image as a graph~\cite{johnson2015image}. 
%The graph is not associated with a specific image but can rather be grounded to an image.
It is defined as a tuple $G=(O,E)$, where the set of vertices  $O$ represents the image objects and the set of edges $E$ represents relationships between objects.
Each object contains the object type and relevant attributes; each edge contains the relationship type.
Scene graphs are used in numerous applications in computer vision, such as image retrieval~\cite{johnson2015image,qi2017online,wang2020cross}, image captioning~\cite{gao2018image,gu2019unpaired,yang2019auto}, VQA~\cite{ghosh2019generating,shi2019explainable} and image synthesis~\cite{johnson2018image,mittal2019interactive,tripathi2019compact}.
A widespread dataset of scene graphs is {\em Visual Genome (VG)}~\cite{krishna2017visual}, which contains scene graphs of 108K images.
Furthermore, there are various algorithms for scene graph generation (SGG), such as%~\cite{han2021image,tang2020unbiased,xu2017scene,yang2018graph,zellers2018neural,zhang2019graphical}.
~\cite{han2021image,tang2020unbiased,xu2017scene,zhang2019graphical}.

%\vspace{0.03in}
\noindent
{\bf Small high-quality datasets and large low-quality datasets.}
Having access to a small trusted dataset, as well as to a large untrusted dataset,
%, which contains label noise and is cheap to assemble.
is a typical scenario in classification with label noise~\cite{hendrycks2018using,li2017learning,veit2017learning,xiao2015learning}.
Another domain is {\em neural machine translation (NMT)}, in which the trusted dataset contains expert translations.
The lack of data in this domain is treated by data mining or other automatic methods that provide new data in varying quality.
This data is considered noisy, therefore used for training jointly with the smaller trusted dataset~\cite{axelrod2017cynical,moore2010intelligent,wang2018denoising,zhang-etal-2019-curriculum}.

%%%%%%%%%%%%%%%%%%%%%%%%%%%
%% section: Method
%%%%%%%%%%%%%%%%%%%%%%%%%%%
\section{Method}
\label{sec:method}
Given an image and a desired length level, our goal is to generate a description that satisfies the length constraint.
Existing captioning models are trained on the same datasets, thus they share the same length statistics.
%Our goal, however, is to generate captions even beyond the length of the available training captions;
We aim to generate descriptions of varying lengths, even when there are very few examples of a certain length.
%beyond the length of the mostly-available (or even any available) training captions; we note that we are the first to address this issue.
Towards this end, in Section~\ref{sec:method} we propose a novel approach that handles the scarcity of long-captions.
We assume that there exists some base model that is capable of generating short captions, for which there is ample data.
We show that our approach manages to preserve this model's performance for short captions, while dramatically improving that performance for long ones.
In Section~\ref{sec:results} we also show that our approach is as beneficial for paragraph generation.

\begin{figure}[tb]
\centering
\begin{tabular}{c}
     \hspace{-0.27cm}
     \includegraphics[width=1.0\linewidth]{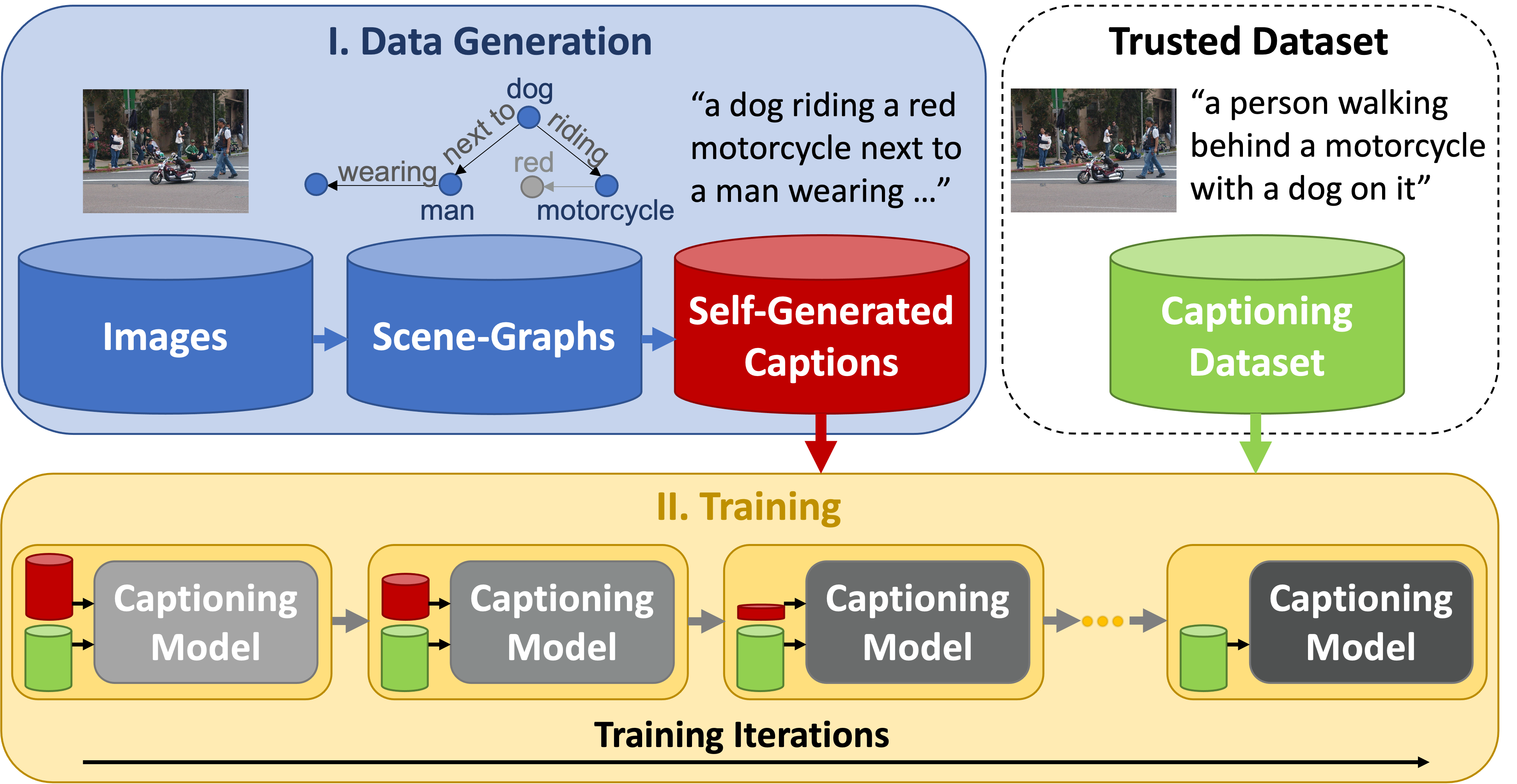}
\end{tabular}
\vspace{-0.1in}
\caption{\textbf{Outline.}
(I) To overcome the shortage in long captions in trusted datasets (\textcolor{greenmethod}{green}), a new dataset is self-generated (\textcolor{redmethod}{red}) using scene graphs, creating an extended dataset.
(II) During training, the low-quality data is gradually filtered out, while remembering the information learned from it.
This improves length control and preserves captioning quality.
}
\label{fig:method}
\vspace{-0.08in}
\end{figure}

Our approach consists of two key ideas, which lead to a two-phase method (Fig.~\ref{fig:method}). 
First, to solve the shortage of long captions, we enrich the dataset with varying-length captions, by utilizing a different image representation, a scene graph.
%At first we will enrich the dataset with varying-length captions (even long), by utilizing other image representations.
% In particular, we will parse scene graphs of images, in order to generate varying-length captions, as explained in Section~\ref{sec:self-generation}.
% This idea, however, solved the problem only partially, since captions generated this way, especially long captions, are prone to many errors, mainly object detection errors and unnatural language style.
%This will solve the length issue.
Second, given a {\em self-generated} dataset, which contains captions of various lengths and varying quality,
%, but might be of low quality and prone to errors, mainly in object detection and in language style.
our novel training procedure benefits from the varying lengths and is barely affected by the varying qualities.

To realize the first phase, any length-aware base model can be utilized.
We use {\em LaBERT}~\cite{deng2020length}, since it is currently the only model that controls the caption length.
This transformer-based model's
%, as illustrated in Fig.~\ref{fig:labert}.
 inputs are visual and word embeddings, where
%the visual embeddings represent image regions and their position in the image and 
%the word embeddings represent the words, their position and the caption length.
the word embeddings also represent the caption length.
This phase is described in Section~\ref{sec:self-generation}.

% figure

For the second phase, 
we propose a training procedure that makes use of the {\em extended dataset}, which consists of the mixed-quality {\em self-generated} dataset and the {\em trusted} dataset.
Our strategy gradually tunes the model, to account for the diverse quality of the data, in a manner in which late iterations are
exposed mostly to high-quality data, while not forgetting valuable information learned from the low-quality data.
This phase is described in Section~\ref{sec:training}.

%%%%%%%%%%%%%%%%%%%%%%%%%%%
%% section: Data Generation
%%%%%%%%%%%%%%%%%%%%%%%%%%%
\subsection{Self-generating varying-length captions}
\label{sec:self-generation}

Our goal is to generate a large dataset of image captions of various lengths, where we are willing to compromise on the quality of the captions.
We hereby describe an algorithm that given an image, will generate such image captions.
It is designed to satisfy three requirements:
(1)~The length of the captions should vary.
(2)~Each caption should include the important objects \& relations.
(3)~The linguistic style should be as natural as possible.

At the base of the algorithm is the use of an image representation comprising the image's objects, their attributes, their importance, and the relationships between them.
This representation is parsed and captions of different lengths are generated, so as to take object importance into account.

% \vspace{0.03in}
\noindent
{\bf Extraction of image information.}
Scene graphs allow us to extract the above information, with the exception of object importance.
Briefly, in this directed graph, the vertices are objects and the edges are relationships.
The objects may contain additional descriptive attributes, such as adjectives. 
%The relationships are either actions or any relation between two objects, such as relative position.
Scene graphs are available for some datasets~\cite{krishna2017visual} or may be automatically generated%~\cite{han2021image,tang2020unbiased,xu2017scene,yang2018graph,zellers2018neural,zhang2019graphical}.
~\cite{han2021image,tang2020unbiased,xu2017scene,zhang2019graphical}.

The only essential information that scene graphs do not hold is objects' importance.
This information is necessary, as captions of all lengths describe the salient objects. 
To gain this information, we compute an image saliency map, utilizing the method of ~\cite{JIA20EML}, which is accurate and easy to run.
% There exists a variety of image saliency detection algorithms~\cite{cornia2018sam,JIA20EML,kruthiventi2017deepfix,Kummerer_2017_ICCV,qi2019convolutional}.
It assigns each pixel a saliency score, which corresponds to human fixation.
The object saliency is computed as the normalized sum of the pixel saliency scores of the object's bounding box.
In our case, the bounding box is provided by the scene graph.
We create a vector whose entries are saliency-based weights of the image objects.

% \vspace{0.03in}
\noindent
{\bf Varying-length caption generation.}
This is the core of our method.
The idea is to explore the scene graph in a {\em Depth First Search (DFS)}-like manner, with a few modifications, which account for caption diversity and length diversity.

The algorithm proceeds as follows.
    The first object is chosen among the nodes of the scene graph, according to the distribution of the saliency vector.
    The more salient an object is, the more likely it is to be chosen.
    Once the initial node (object) is selected, DFS is applied with three differences.
    First,  we do not necessarily visit all the children of each node.
    Instead, we sample up to $k$ children, based on their saliency distribution (after re-normalization). 
    % of its children re-normalized (as it is only a subset of vertices.
    % Thus, the order of the visited children is drawn from the saliency-driven distribution.
    Furthermore, we allow re-visiting nodes, since captioning "re-visits" objects, by mentioning them multiple times.
    For example, the object "boy" in the description "A boy is feeding a dog and a girl is smiling at the boy", is revisited, where the second time is preceded by "the". 
    While revisiting nodes is allowed, loops are prohibited, in order to prevent (infinite) repetitions. % of the same objects and relationships.
    This is done by avoiding propagation through already-visited vertices.
    Finally, if the exploration reaches an end, but less than $T_{sal}\%$ of the sum of the image saliency map was explored, we "jump" to a new unexplored node, selected again according to the saliency vector. 
    This means that we do not necessarily visit the whole graph.

    During this traversal, a caption is generated, creating a noun for every node (object), a verb for every edge (relationship) and adjectives for the objects (node attributes).
    As a common connection of phrases in captioning, we add an "and" when there is a jump in the DFS (or a dot in the case of paragraphs).
   To increase the diversity of the captions and their lengths we add two random procedures:
    First, we cut out a random number of the last visited objects. 
  Second, for every object in the generated caption we add a random number of  attributes (up to a limit $n_a$).
The algorithm is summarized as pseudo-code in the supplemental materials.

% figure
\begin{figure}[tb]
\centering
\begin{tabular}{c}
     \includegraphics[width=0.7\linewidth]{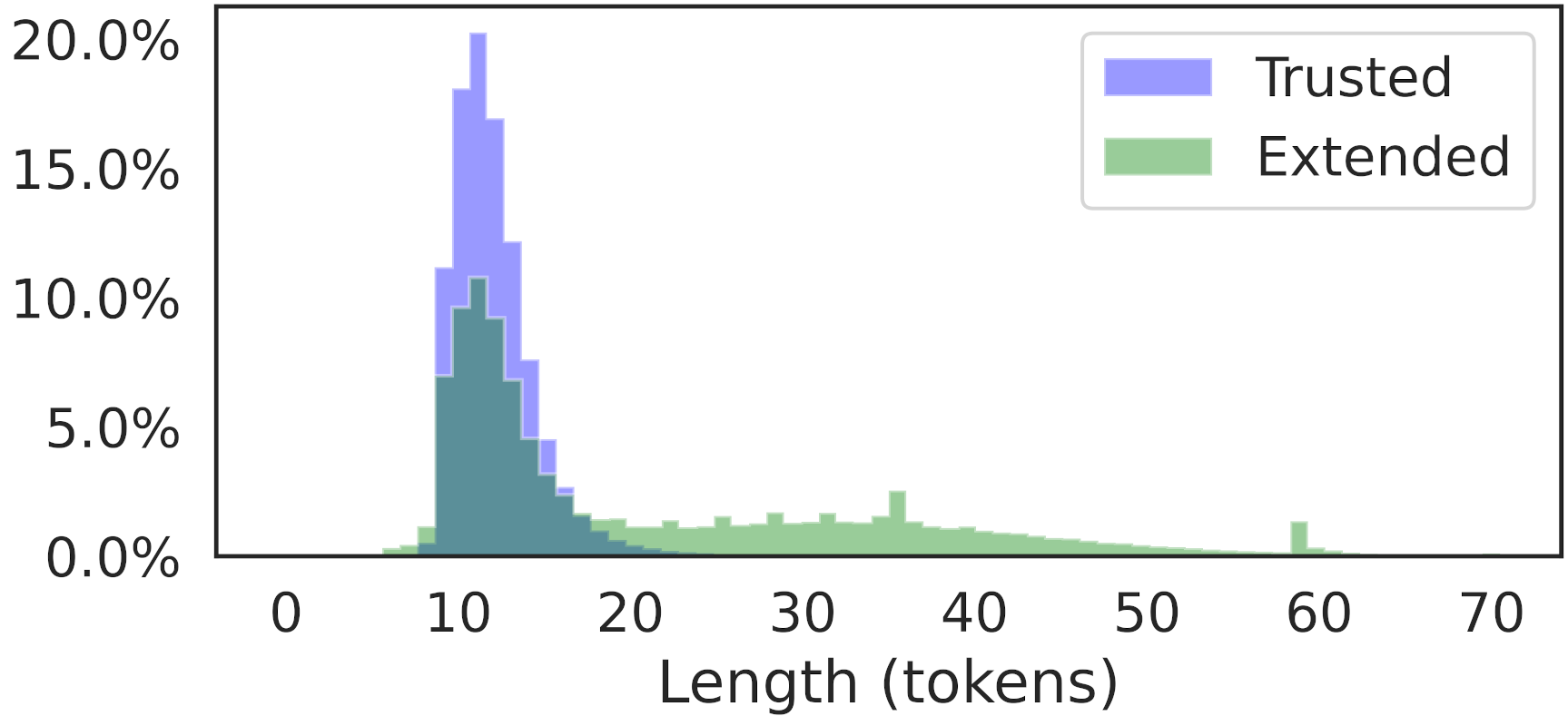}
\end{tabular}
\vspace{-0.08in}
\caption{\textbf{Length of captioning datasets.}
The average caption length in the trusted (MS-COCO) dataset is $11.95$ tokens with standard deviation of $2.58$, whereas in our extended dataset these are $21.3$ and $13.56$, respectively.
%Thus, the extended dataset contains longer captions.
(Overlaps cause the third color.) 
}
\label{fig:data_stat}
%\vspace{-0.2in}
\end{figure}

% \vspace{0.03in}
\noindent
{\bf Linguistic style.}
The generation algorithm described above successfully depicts the objects, their order, and the relationship between them.
However, it might generate unnatural linguistic style.
A notable example
%, which is fixed in our method, 
is {\em grouping:} multiple objects (e.g. "two kids" or 
% objects that form a group (e.g.
"animals") appear in scene graphs as distinct nodes and are thus  described as separate objects.
To  improve the grammar and the style,
%and minimize the domain gap, 
we use an off-the-shelf language paraphraser~\cite{huggingface2021pegasus}, which
% to make our parsed captions into more natural captions.
preserves the original meaning 
% (high adequacy) 
and produces a fluent and correct language.
%(high fluency).
The paraphraser itself has no access to the image.
% Note that this is a text-to-text conversion, with no visual constraints.
% , which is pre-trained on large corpora.
Note that the paraphraser is applied only here and not on the output of the model described in Section~\ref{sec:training}.

% \vspace{0.03in}
\noindent
{\bf Results.}
Fig.~\ref{fig:data_stat} compares the  datasets.
The captions in the original {\em trusted} dataset are relatively short, where $98\%$ of the captions have less than $20$ tokens.
Differently, in our {\em extended} dataset only $53\%$ are shorter than $20$ tokens and the number of long captions is no longer negligible.
We set $T_{sal}=80\%$, $k=2$ and $n_a=4$, which suffice to explore both the essence of the image and many details.

%%%%%%%%%%%%%%%%%%%%%%%%%%%
%% section: Data Selection
%%%%%%%%%%%%%%%%%%%%%%%%%%%
\subsection{Training with data selection}
\label{sec:training}

% Deep models are eager for data, which is often a pre-requisite for achieving high accuracy.
% Our domain is not the only one where  high-quality data for supervised training is expensive or out of reach.

We are given a small high-quality  {\em (trusted)} dataset  and a large {\em self-generated} dataset.
Our key assumption is that although the self-generated dataset is of low-quality, it does contain beneficial information for training.
However, it should be utilized thoughtfully, in order not to be harmful.
%A similar assumption was made in machine translation of~\cite{wang2018denoising}, where  the trusted dataset contains expert translations, while the noisy dataset is automatically generated.
%The training procedure is therefore a continuous tuning of the model, while filtering the dataset differently at each stage.

A similar challenge is addressed in other domains, such as
% Careful use of low-quality data, which is easier to collect, has opened new opportunities in other domains, for instance in 
 training classifiers with noisy labels~\cite{liu2020peer,xiao2015learning,zhang2018generalized} or using synthetic data~\cite{chen2019learning,mu2020learning,sankaranarayanan2018learning}. 
%~\cite{carlson2018modeling,chen2019learning,dahmen2019synsys,mu2020learning,richardson20163d,sankaranarayanan2018learning,shermeyer2021rareplanes}.
 %,tremblay2018training,varol2017learning}.
The most related domain to ours is machine translation (NMT), where expert translations are rare, while noisy translations can be automatically generated.
Our work, which is the first to explore mixed-quality datasets for image description, is inspired by~\cite{wang2018denoising}.

% Our core idea is to thoughtfully use low-quality data, which we generate, as explained in Section~\ref{sec:self-generation}.
%
We propose a training strategy that gradually tunes the model, to account for the diverse quality of the data.
The core idea is to expose the model to the diverse data early on, and to expose it  mostly to high-quality data  in late stages of the training.
% As the quality of the captions varies, we shall select the data wisely.
This is so since the low-quality data provides a mass of fundamental information (objects, attributes, length control signals)  that is essential for learning.
The high-quality data is important later, since it fine-tunes the model on more relevant domain data. 
The benefit of the entire data is supported by our experiments in Section~\ref{sec:results}.

To gradually filter out  data, we shall find a measure that rates the data according to its quality.
Intuitively, the quality of a data point can be determined by its domain relevance, measured by the distance to in-domain data.
% Then, we can filter the data according to these rates as the training progresses.
Similarly to~\cite{wang2018denoising}, this is done by
the ratio between the probability to appear in the trusted training dataset to the probability to appear in the extended (trusted \& self-generated) dataset.
The higher the ratio, the more likely the data point to contain high-quality information to learn from.

Formally, assume for now that a captioning model, parameterized by $\theta$, can output the probability $p(y_j|x_j;\theta)$ of a caption $y_j$ to match an image $x_j$ (this assumption will be later removed). 
%(It will be explained below, in Eq.~\ref{eq:prob}).
Given two models:  $M_\theta$, which is trained on the trusted dataset $D$, and  $M_{\Tilde{\theta}}$, which is trained on the extended dataset $\Tilde{D}$, we approximate the quality of each data point $j$ in the self-generated dataset by:
\begin{equation}
    \begin{split}
    u_j & = quality(x_j, y_j ; \theta, \Tilde{\theta}) = \log \left( \frac{p(y_j|x_j;\theta)}{p(y_j|x_j;\Tilde{\theta})} \right) = \\ 
    & = \log p(y_j|x_j;\theta) - \log p(y_j|x_j;\Tilde{\theta}).
    \end{split}
    \label{eq:noise}
\end{equation}
By Eq.~\ref{eq:noise}, a positive value means that a data point is more likely according to the model trained on the trusted data than that trained on the extended data, while a negative value means the opposite.
Therefore, this score rates the data points according to their likelihood of being in-domain.

\begin{figure}[tb]
\centering
\begin{tabular}{c}
    % \hspace{-0.5cm}
    %  \raisebox{-0.05cm}{\includegraphics[width=0.5\linewidth]{figs/percentiles/coco_percentiles}} &
    %  \hspace{-0.25cm}\includegraphics[width=0.53\linewidth]{figs/percentiles/sep_box} \\
    %  (a) Quality percentiles & (b) Quality by length level
    \includegraphics[width=0.67\linewidth]{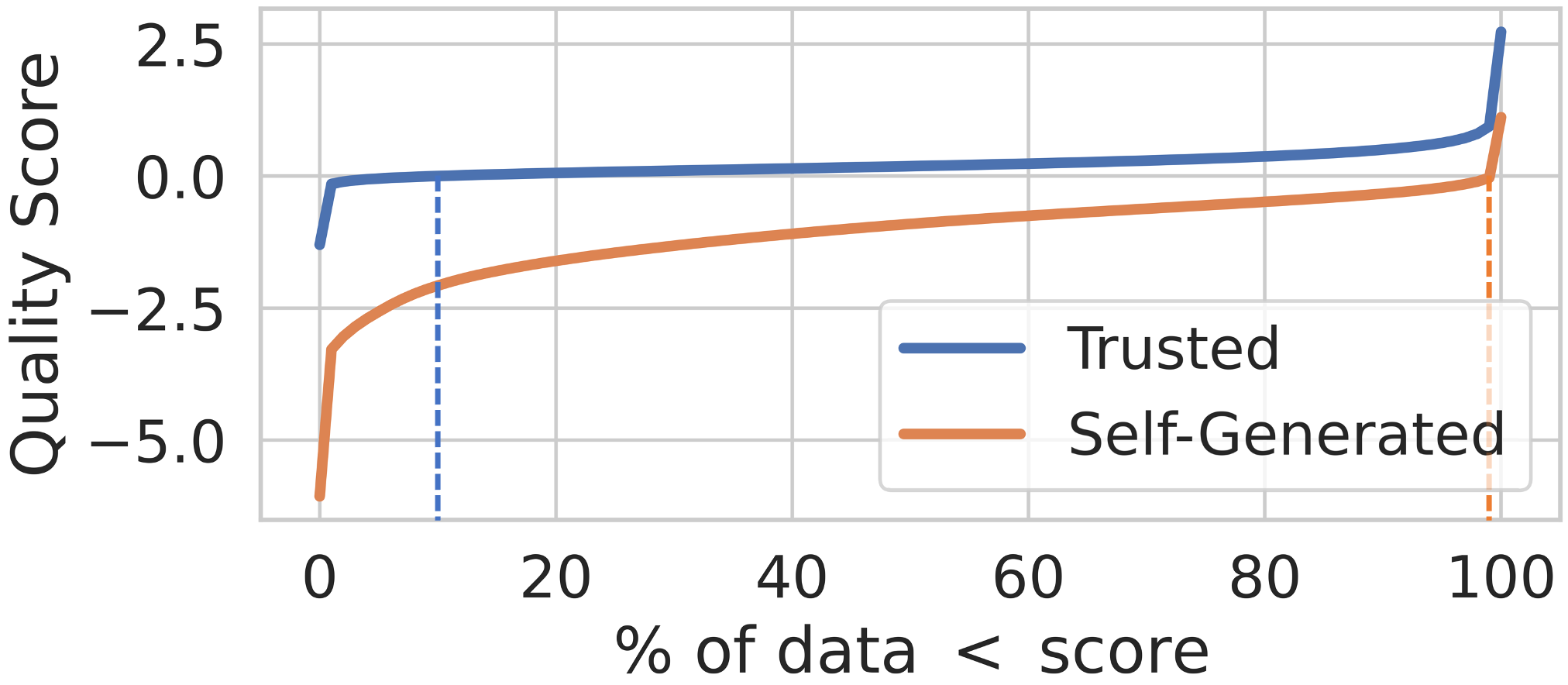}
\end{tabular}
\vspace{-0.08in}
\caption{\textbf{Data separation.}
Quality scores, computed by Eq.~\ref{eq:noise}, manage to separate the trusted data from the self-generated data.
%(a) the percentiles for both datasets across length levels.
The trusted data points have mostly ($90\%$) positive scores 
%($10\%$ negative; 
(\textcolor{bluedashed}{dashed blue} line), 
while the self-generated data points have mostly ($99\%$) negative scores (\textcolor{orangedashed}{dashed orange} line).
}
\label{fig:percentiles}
%\vspace{-0.05in}
\end{figure}

%\elad{Delete:} Fig.~\ref{fig:percentiles} shows that indeed this approach manages to softly \ayellet{what is softly?} separate the trusted data from the noisy data on MS-COCO dataset.
Fig.~\ref{fig:percentiles} shows that our approach successfully separates most of the trusted data from the self-generated data on MS-COCO.
%Fig.~\ref{fig:percentiles}(a) shows that 
While $90\%$ of the trusted data has a positive score (high quality), it holds for only $1\%$ of the self-generated data.
% \elad{A per-level separation reveals that the low-levels (short captions) of the self-generated data are of overall higher quality, even containing captions with positive scores, while higher-levels (long captions) are of a lower quality.
% In the trusted dataset, however, the captions (also long) mostly have a higher \& positive score.
% }

% figure

Luckily, $p(y_j|x_j;\theta)$ can be estimated by language models that learn to predict tokens.
We extend~\cite{vaswani2017attention} to consider the additional input image, as follows.
% The transformer model of~\cite{vaswani2017attention} predicts the probability of every token to appear in masked input locations.
%The model is trained by masking input tokens and predicting them by those that achieve the maximal probability according to the model.
%In particular, 
Given a caption $y_j$, which is a sequence of $m$ tokens $[w_1,...,w_m]$, and an image $x_j$, the likelihood of the caption depends on the likelihood of each token to appear, based on the previous tokens and the image.
For a token $w_t$ the probability is $P _\theta (w_t | w_{1:t-1}, x_j)$ and the log-likelihood of the caption is
\begin{equation}
    \log p(y_j|x_j;\theta) = \frac{1}{m} \sum _{t=1} ^m \log \left( P _\theta (w_t | w_{1:t-1}, x_j) \right).
    \label{eq:prob}
\end{equation}

%Hereafter we proceed to describe our data selection algorithm.
\vspace{-0.02in}
Our data selection algorithm uses this estimation
%The basic idea is to iteratively  compute a {\em quality threshold} and filter out the data points below that threshold.
 to gradually filter out data points below a {\em quality threshold}.
As the training proceeds, the threshold increases, eliminating more and more self-generated data points. 
Fig.~\ref{fig:percentiles} shows that low-quality data is filtered at early iterations, while higher-quality data remains until later iterations.

A unique challenge that differentiates our length-control task from other tasks that use diverse-quality datasets is that
% , of length control, from NMT is that
%that of machine translation~\cite{wang2018denoising}.
% an inherent property of our self-generated data is that 
almost all long captions tend to have low-quality scores.
Therefore, long captions are prone to catastrophic forgetting, as they are eliminated early on.
This could reduce the control abilities for long captions.
To avoid this situation, we make sure that low-quality captions (e.g., the long ones) will appear in late training iterations.
%This is crucial as otherwise the long captions would be filtered out very early.
This, however, might contradict the core idea of our training, which requires only high-quality captions to appear at late training stages.
We propose to allow a small number of low-quality self-generated 
captions to appear  as the training progresses, by adding some randomness, as follows.

Without adding the above randomness, we may view our filtering procedure at the $i^{th}$ iteration as sampling according to a probability function, where data points with a quality score $<T_i$ get probability $0$ and all other points get the same positive probability.
% Filtering $N$ data points whose quality score is greater than a threshold $T$ is equivalent to assigning probability $0$ to any data point whose quality is smaller than $T$ and a positive probability otherwise. 
% % Then, sampling $N$ points according to these probabilities (without replacement). 
% Hence, we \ayellet{?model} data selection as sampling according to a probability function, where  data points with a quality score $<T$ get probability $0$ and all other points get the same positive probability.
This function is essentially a step function, centered at $T_i$, which maps quality scores to probabilities.
%When all data points with a quality score $<T$ have probability $p=0$ and all other points have an equal probability $p>0$, the probability function is in essence a step function centered at $T$, which maps a quality score to probability.
%
% With a small change to this model of selection, we can integrate low-quality data points wisely. 
% During training, $<T$ increases and \ayellet{?hedges} the selected data quality.
We  add randomness by smoothing the step function %(Fig.~\ref{fig:smoothstep})
%Recall that we wish to soften this threshold and allow a small number of lower-quality data points to be selected. softening this threshold 
and allowing a small number of low-quality data points to be selected accordingly. 
% This is done by smoothing the step function.
%, which is used for assigning the probabilities, to achieve that.

% table
\begin{table*}[htb]
    \centering
    \begin{tabular}{l|ccccccc|c}
        \toprule
        Level &  1 &  2 &  3 &  4 &  5 &  6 & 7 & Average \\
        Length (tokens) & 1-9 & 10-19 & 20-29 & 30-39 & 40-49 & 50-59 & 60-69 &\\
        \midrule
         \% in trusted dataset & 11.8\% & 86.6\% & 1.4\% & 0.08\% & 0.02\% & 0.004\% & 0\% & \\
         \% in extended dataset & 9.1\% & 51.8\% & 13.1\% & 13.5\% & 7.7\% & 3.8\% & 0.1\% & \\
        \midrule
        LaBERT~\cite{deng2020length} &  \textbf{100\%} & 98.03\% & \textbf{93.75\%} & 83.78\% & 49.92\% & 51.19\% & 0\% & 68.10\%\\
        CLID (ours) & \textbf{100\%} & \textbf{98.64\%} & 92.6\% & \textbf{84.17\%} & \textbf{76.99\%} & \textbf{69.18\%} & \textbf{22.13\%} & \textbf{77.67\%}\\
        \bottomrule
    \end{tabular}
    \vspace{-0.05in}
    \caption{\textbf{Captioning control precision}. 
    For each level, the second row shows the range of tokens for this level.
    The next two rows show the percentage of captions of each level in the training datasets.
    The two bottom rows compare the precision results of~\cite{deng2020length}
    %, CLID's training strategy without data selection but rather only pretraining with our extended data 
%    \ayellet{isn't that Deng et al.+extended?} \elad{It's pretraining instead of gradual filtering}  (only Section~\ref{sec:self-generation}) and  
to ours. %(sections~\ref{sec:self-generation} and~\ref{sec:training}).
    We outperform ~\cite{deng2020length} both on average and  for most of the levels, with the exception of Level $3$ (which is a prevalent level in the trusted dataset).
    As expected, our benefit is mostly evident for the high levels.
    The results are the mean of $3$ independent runs.
%    \elad{Each model should generate captions from all levels. The average score is therefore across all levels.}
    }
    \label{tab:control_precision}
    \vspace{-0.1in}
\end{table*}

In particular, given a threshold, $T_i$, we create
% Our selection process for training stage, $i$, is performed by first setting a threshold, $T_i$, to this stage, then creating 
a smooth step function in accordance with $T_i$ and perform data selection, as follows.
The caption quality $u_j$ of a data point $j$ (Eq.~\ref{eq:noise}) and $T_i$ are used to compute the probability function needed for data selection.
We use the following smooth step function, centered at the threshold $T_i$:
%\elad{Delete: }where the value of $T$ is the $p_i^{th}$ percentile, matching the training phase $i$: \ayellet{unclear}
\begin{equation}
    f(u_j;T_i,s) = \frac{1}{2} \cdot \left[ 1 + \tanh \left( \frac{u_j-T_i}{s} \right) \right].
    \label{eq:step}
\end{equation}

Here, $s$ is a  tunable parameter, which controls the smoothness of the function.
%
%Fig.~\ref{fig:smoothstep} demonstrates the effect of $s$.
For $s \rightarrow 0$, we get an ideal step function, which means that values smaller than $T_i$ have $0$ probability to be sampled.
The larger $s$ is, the greater the probability of low-quality data points to be selected.
Since self-generated long captions are usually assigned low quality scores (Fig.~\ref{fig:percentiles}), a small $s$ might cause long captions to be eliminated early on in the training process.
Reversely, if $s$ is too high, too much of out-of-distribution data will appear at late training iterations.
We note that an additional benefit of our scheme is that it controls all low-quality captions and not only long ones.

\vspace{0.03in}
\noindent
{\bf Training---putting it all together.}
% Algorithm~\ref{alg:training} summarizes our proposed training procedure.
Hereafter we recap our training procedure with data selection.
We are given a trusted dataset $D$, an extended dataset~$\Tilde{D}$,
%(trusted \& self-generated)  
 and a step smoothness value $s$. 
Two captioning models are trained, one for the trusted dataset and the other for the extended dataset, $M_\theta$ and $M_{\Tilde{\theta}}$, respectively.
The quality score~ $u_j$ of each data point in the self-generated dataset is estimated using the two models (Eq.~\ref{eq:noise}).
%this score will later be used in the data selection process.
Then, the captioning model is trained with iterative data selection until all the self-generated data is filtered out, as follows.
At iteration $i$,
\begin{enumerate}
    % \vspace{-0.1in}
    \item 
    A new threshold $T_i$ is computed. 
    Specifically, it is determined by the amount of the generated data to be removed, according to the $c \cdot i$ percentile.
    %($c=2$ in our experiments).
    For instance, at the $5^{th}$ iteration and for  $c=2\%$, $T_5$ is set such that $10\%$ of the data will be below the threshold.
    \vspace{-0.1in}
    \item
    A weight for each data point, $f(u_j;T_i,s)$, is computed, depending on its quality score, the threshold and the smoothness value (Eq.~\ref{eq:step}).
\vspace{-0.1in}
    \item
    The self-generated dataset is randomly sampled, based on these weights.
\vspace{-0.1in}
    \item
    The model is trained on the trusted \& the sampled self-generated datasets (for duration $\eta$).
    % In our implementation, the duration of each iteration is set to a pre-defined value $\eta$ (=$1200$ iteration in our experiments).
\end{enumerate}
\vspace{-0.08in}
% 
% The model and the pseudo-code are given in the supplementary materials.
See the supplementals for the model and pseudo-code.

\begin{figure*}[htb]
\centering
\begin{tabular}{ccc}
     \hspace{-0.1cm}\includegraphics[width=0.3\linewidth]{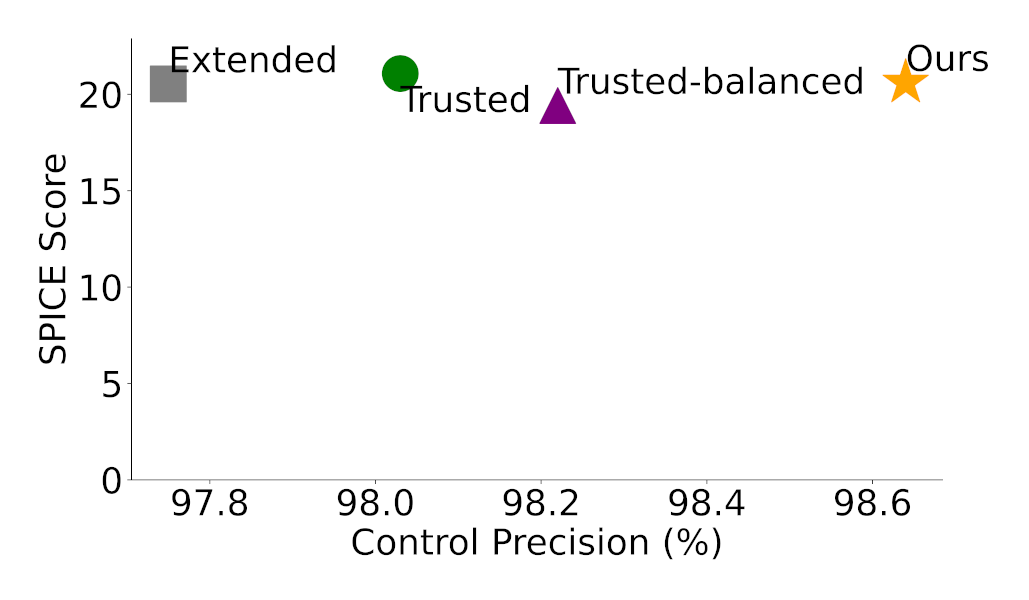} &
     \hspace{-0.1cm}\includegraphics[width=0.3\linewidth]{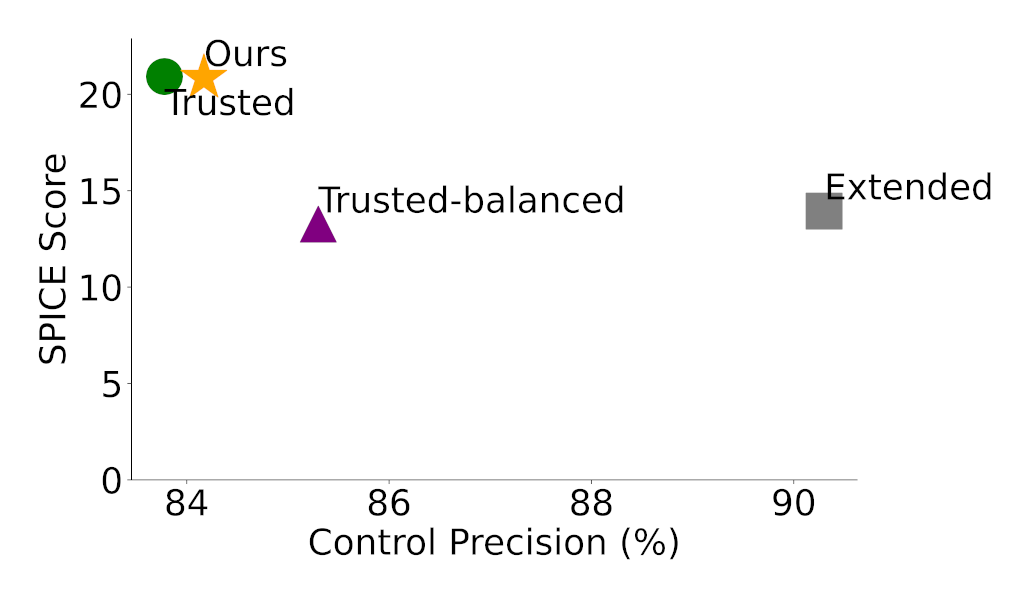} &
     \hspace{-0.1cm}\includegraphics[width=0.3\linewidth]{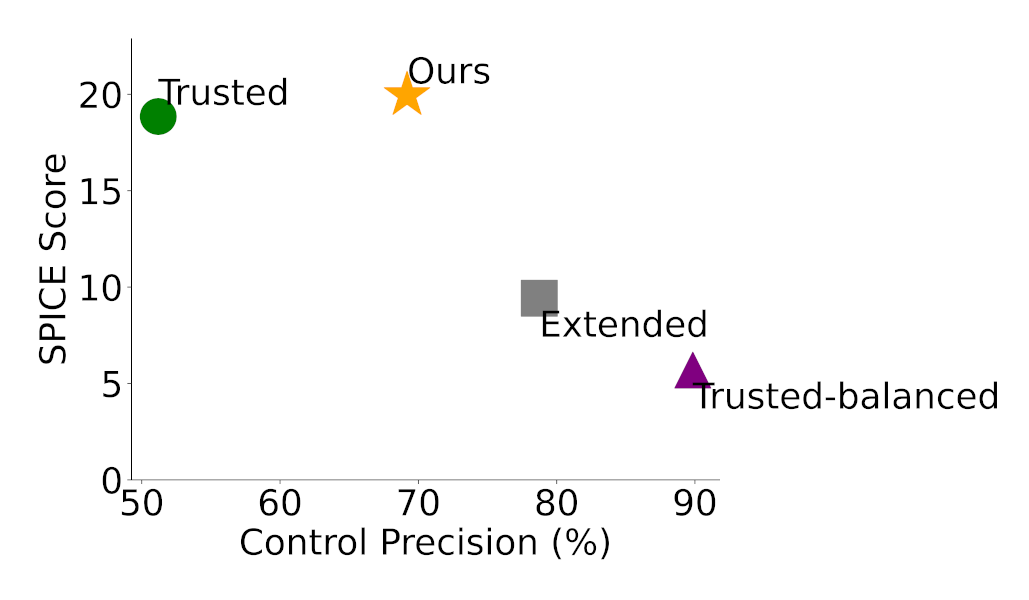} \\[-0.08in]
     % \vspace{-0.15in}
     (a) Level 2 & (b) Level 4 & (c) Level 6 \\
\end{tabular}
\vspace{-0.1in}
\caption{\textbf{Captioning performance.}
In terms of the SPICE quality  measure (vertical axis), our results (\textcolor{orangescatter}{orange star}) are similar to 
~\cite{deng2020length}'s (\textcolor{greenscatter}{green circle}), which is trained on the trusted dataset.
The quality of other solutions (\textcolor{grayscatter}{gray}/\textcolor{purplescatter}{purple}) is dramatically degraded.
While comparable to~\cite{deng2020length} quality-wise, our model  improves the control precision (horizontal axis).
In both measures, higher is better.
The figure shows $3$ length levels; the other levels appear in the supplements.
}
%While other solutions, such as balancing the trusted dataset or training with the extended dataset degrade the output quality (lower SPICE scores), our method achieves similar or better SPICE scores than training with the trusted dataset and improves the control precision.
%The full figure (levels 1-7) appears in the supplementals.
\label{fig:spice}
%\vspace{-0.08in}
\end{figure*}

%%%%%%%%%%%%%%%%%%%%%%%%%%%
%% section: Experiments
%%%%%%%%%%%%%%%%%%%%%%%%%%%
\section{Experimental results}
\label{sec:results}

%This section presents our results and an ablation study.

%%%%%%%%%%%%%%%%%%%%%%%%%%%%%%%%%%%%%
%%%% Datasets and Metrics
%%%%%%%%%%%%%%%%%%%%%%%%%%%%%%%%%%%%%
\vspace{0.03in}
\noindent
{\bf Datasets.}
We use  {\em MS-COCO Caption}~\cite{chen2015microsoft}, which is the most common captioning dataset in general, and the only dataset used in a previous length-control work~\cite{deng2020length}.
%
%MS-COCO contains $123,287$ images with $5$ captions each.
It contains $5$ captions per image, with $113,287$ training images, $5000$ validation images and $5000$ testing images, according to the widely-used Karpathy's split~\cite{karpathy2015deep}.
As seen in Table~\ref{tab:control_precision}, Level 2 contains the majority of the captions ($87\%$). %(87\% \& 52\% respectively).
There are hardly any longer captions and no captions in level $7$.

The {\em Visual Genome (VG)}~\cite{krishna2017visual} dataset contains scene graphs for images.
%, for MS-COCO training images  whenever possible.
Its intersection with MS-COCO is around $50\%$.
For the other MS-COCO images, we use a scene graph generator~\cite{han2021image}.
% , therefore we use VG scene graphs in that case.

{\em Flickr-30k}~\cite{plummer2015flickr30k} is evaluated in the supplementary material.
For this dataset, we utilize only automatically generated scene graphs.

% \vspace{0.03in}
\noindent
{\bf Evaluation metrics.}
In addition to checking whether a description meets the length constraint, we measure its quality.
% We measure both the captioning quality and whether it satisfied the length constraint.
%
We use the {\em control precision} metric~\cite{deng2020length} to measure the percentage of descriptions of the desired length.
There are several possible NLP metrics for measuring quality that calculate the similarity between a generated description and the ground truth, including BLEU~\cite{papineni2002bleu}, ROUGE~\cite{lin2004rouge}, METEOR~\cite{banerjee2005meteor}, CIDEr~\cite{vedantam2015cider}, and SPICE~\cite{anderson2016spice}.
As in~\cite{deng2020length}, we focus on SPICE, which is 
% based on the similarity between scene graphs that are generated from the ground-truth and the output captions. It is 
more robust to caption length and has a high correlation with human judgement.
% Most of these metrics (all but SPICE) are based on $n$-grams.
% Due to the relatively short captions in the dataset (both in the training and test sets), such metrics are expected to degrade when the captions are long.

%\subsection{Results}
%\label{subsec:results}

\noindent
{\bf Length-control results.}
Table~\ref{tab:control_precision} demonstrates the effectiveness of our approach in terms of control.
The control precision is excellent for captions of Levels $1$-$3$ and then starts dropping.
Our approach significantly outperforms the average results of~\cite{deng2020length}.
Specifically, it improves the results in all levels, except for Level $3$, and vastly improves the results of the high levels.

% figure
\begin{figure*}[tb]    
    %\vspace{0.2cm}
    \begin{minipage}[b]{0.15\linewidth}
    \hspace*{0.2cm}
    \raisebox{-0.28\height}{\includegraphics[width=1.33\linewidth]{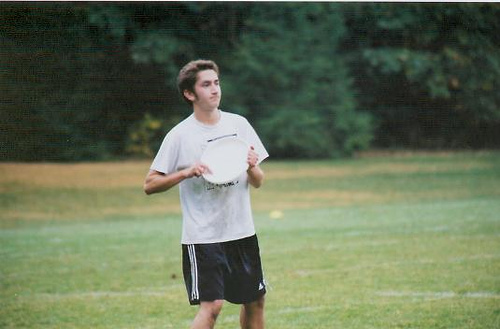}}
    \end{minipage}
    \begin{minipage}[b]{0.85\linewidth}
    \begin{flushright}
    \resizebox{0.98\textwidth}{!}{%
        \hspace{1cm}
        \begin{tabular}{m{0.08\linewidth} m{0.1\linewidth}  m{0.82\linewidth}}
         \toprule
         \multirow{2}{*}{\raisebox{-1.5cm}{\shortstack[l]{LaBERT \\[0.02in] \cite{deng2020length}}}} & Level $2$ &  Two men in a white t-shirt are playing frisbee. \\
          \cmidrule{2-3}
         & Level $5$ & A man is playing a park. He is reaching for a frisbee. The man is wearing a white shirt with white stripes down the side. The frisbee is white and has a lot of grass on it. \\
         \cmidrule{2-3}
         & Level $8$ & Two men are playing frisbee on a grassy field. Both of the men are wearing black t-shirts with white stripes down the side of the shirts. One man is holding a frisbee in his hand. The field is covered in green grass. Part of green trees can be seen behind the men. It appears to be daylight outside. \\
         \midrule[0.8pt]
         \multirow{2}{*}{\raisebox{-1.8cm}{{\shortstack[l]{CLID \\ (ours)}}}} & Level $2$ & A man in a white shirt and black shorts holds a frisbee. \\
         \cmidrule{2-3}
         & Level $5$ & A man is standing outside on a grassy field. He is holding a white frisbee in his hand. The man
            is wearing a white t-shirt and black shorts. There are trees behind the man in the background. \\
         \cmidrule{2-3}
         & Level $8$ & A man is playing frisbee on a field. He is holding a frisbee in his hand. The man is wearing black and white shirt and shorts. The man has black and white stripes on the shorts. He has dark hair. The field is green with green grass behind the trees in the field. The man is standing in the field. \\
         \midrule[0.8pt]
         & (a) Length & (b) Generated Caption
        \end{tabular}
        }
    \end{flushright}
    \end{minipage}
    
\vspace{-0.1in}
\caption{\textbf{Qualitative comparison (paragraphs).}
Comparing our outputs to those of~\cite{deng2020length} demonstrates similar quality for the short descriptions.
However, our long descriptions are more coherent and detailed than~\cite{deng2020length}'s. 
%See more examples in the supplementals.
}
\label{fig:qualitative}
%\vspace{-0.1in}
\end{figure*}

%%%%%%%%%%%%%%%%%%%%%%%%%%%%%%%%%%%%
%%%%%%%%% accuracy
%%%%%%%%%%%%%%%%%%%%%%%%%%%%%%%%%%%%
% \vspace{0.03in}
\noindent
{\bf Quantitative results.}
%As a result from balancing MS-COCO training set, for example, the quality of the model trained with the balanced data degrades.
Fig.~\ref{fig:spice} shows that our method generates good-quality results, yet with higher control.
%The horizontal axis is the control precision and the vertical axis is the quality measured by SPICE.
The quality of our captions (orange) is comparable to those of~\cite{deng2020length}, trained on the trusted data (green).
% For both measures, higher values are better, hence the optimal region is the top right.
It is also compared to the quality of the results of two other possible solutions.
In the first the model is trained on a length-balanced version of the trusted dataset (purple), and in
% , by balancing the amount of captions at each level (purple).
the second on the extended dataset (gray).
%(this is done with a weighted rather uniform data sampler).
These approaches have high control precision, but on the expense of quality.
The supplemental material extends Fig.~\ref{fig:spice} to all the levels and exhibits similar results.
It also shows results of the other metrics, demonstrating similar improvements.
Furthermore, it includes additional results on Flickr-30k that exhibit a similar behavior.
% again much better overall results for lengthy captions and comparable results for the short captions for most  evaluation metrics.
% and  much better results than other solutions.

In order to demonstrate the impact of high-quality information from short captions on long caption generation, we observe Level $7$, which is absent from the trusted captioning dataset. 
%(60-69 tokens), 
%When our model is trained on the extended data, this level has control precision of 97.96\% but the generated captions achieve an average of 11.49 SPICE score.
When training a captioning model on the extended dataset without any modifications, the average SPICE score for captions at this level is $11.49$.
However, by employing our training procedure, which does not involve the addition of any ground-truth caption at this length level, the same model achieves a significantly improved score of $19.54$.

% \vspace{0.03in}
\noindent
{\bf Qualitative results.}
Fig.~\ref{fig:teaser} illustrates representative outputs generated by our model. 
Notably, our descriptions for higher levels demonstrate greater elaboration and encompass more scene-specific details compared to the descriptions produced by~\cite{deng2020length}. However, short descriptions are comparable for both models.
See supplementary materials for more qualitative results.

To assess the quality of our captions, as rated by humans, we conducted an experiment on Amazon Mechanical Turk. 
The experiment involved presenting a triplet consisting of an image and two captions: one generated by \cite{deng2020length} and one generated by our model. 
Workers were asked to determine which caption better described the image, or if both captions were equally good. 
We randomly selected 300 image-caption triplets and assigned each to three different workers.

Table~\ref{table:userstudy} presents the percentage of worker votes for each model, per caption length (ranging from $1$ to $6$, the levels which both models can generate caption in). 
The table reveals that our captions were preferred by human evaluators across all length levels. 
For instance, in level 1, 60\% of workers preferred our captions, while in level 6, 57\% of workers favored our captions.
An interesting observation is that even though our focus was not primarily on the lower length levels (levels $1$-$3$) due to the abundance of available data, the human evaluation demonstrated a quality advantage in these levels as well.

%%%%%%% table
\begin{table}[t]
\centering
% \small
\begin{tabular}{m{0.83in} | >{\centering}m{0.19in} >{\centering}m{0.16in} >{\centering}m{0.19in} >{\centering}m{0.19in} >{\centering}m{0.19in} >{\centering\arraybackslash}m{0.19in}}
\multirow{2}{*}{Method} 
& \multicolumn{6}{c}{Level} \\
& 1 & 2 & 3 & 4 & 5 & 6 \\ \hline
LaBERT~\cite{deng2020length} & 40\%  & 44\%  & 39\%  & 47\%  & 41\%  & 43\%   \\
CLID (ours) & \textbf{60\%}  & \textbf{56\%}   & \textbf{61\%}   & \textbf{53\%}  & \textbf{59\%} & \textbf{57\%} \\
\end{tabular}
\vspace{-0.1in}
\caption{
\textbf{User study results.}
Percentage of votes for captions generated by each method, per length level.
Human evaluators tend to prefer our captions over those of~\cite{deng2020length}.}
\label{table:userstudy}
\vspace{-0.1in}
\end{table}

\noindent
{\bf Paragraph generation.}
As our proposed method is general, we address the task of paragraph generation similarly to captioning.
A paragraph consists of consecutive sentences separated by dots.
We consider them as a sequence of text tokens, as in captioning.
To create the self-generated dataset, we used the algorithm described in Section~\ref{sec:self-generation}.
The only difference is that a dot was added instead of an "and" when jumping during the DFS traversal.

We use the {\em Stanford Descriptive Paragraphs}~\cite{krause2016paragraphs} dataset, which is the only available dataset for the task.
The images in this dataset are taken from VG.
It contains $14,575$ training images, $2,487$ validation images and $2,489$ test images.
Each image is associated with a single paragraph.
Hence, this dataset is much smaller than the captioning dataset (by a factor of $30$).
$2\%$ of the paragraphs have less than $29$ tokens;
$32\%$  have $30$-$59$ tokens;
$52\%$  have $60$-$89$ tokens; 
and $14\%$  have $90$-$129$ tokens.
%As we are the first to consider length-control paragraph generation, we had to set the length levels.
% We set $13$ uniform length levels, which range from $1$-$129$ tokens. 
% (e.g. level 10 is for 90-99 tokens).
%To create the self-generated dataset, we used the algorithm described in Section~\ref{sec:self-generation}.
%The only difference is that a dot was added instead of an "and" when jumping during the DFS traversal.
We set $13$ uniform length levels, which range from $1$-$129$ tokens. 

Since there are no works on length-control paragraph generation, we trained~\cite{deng2020length} to generate paragraphs, in order to allow comparisons.
Our method improves the control abilities at all length levels, compared to training only on the trusted data.
It achieves an average of $94\%$ in control precision, which is a $14\%$ gain.
In terms of the the common NLP scores (e.g., BLEU),
% (SPICE does not suit paragraph descriptions
%which was used in~\cite{krause2016paragraphs} to measure the quality of paragraphs, 
our model achieves comparable results to~\cite{deng2020length}'s across all levels.
Additionally, Fig.~\ref{fig:qualitative} shows typical outputs of our model.
While our short descriptions are comparable to those of~\cite{deng2020length}, our long descriptions are more coherent and detailed.
Please refer to the supplemental materials for further details and examples.

%%%%%%%%%%%%%%%%%%%%%%%%%%
%%%% section: ablation
%%%%%%%%%%%%%%%%%%%%%%%%%%
\vspace{-0.1in}
\section{Ablation study}

\noindent
{\bf Graph coverage.}
The amount of information extracted from a scene graph is determined by $T_{sal}$ (\%).
A low value of $T_{sal}$ results in overall short generated description datasets.
Table~\ref{table:t_sal} illustrates the impact of this parameter on the model's control ability.
While the control ability is comparable for levels 1-4, which are prevalent in all of these self-generated datasets, the advantage of using larger $T_{sal}$ values becomes pronounced in longer levels.

%%%%%%% table
\begin{table}[t]
\centering
\small
\begin{tabular}{m{0.2in} | >{\centering}m{0.16in} >{\centering}m{0.25in} >{\centering}m{0.25in} >{\centering}m{0.25in} >{\centering}m{0.25in} >{\centering}m{0.25in} >{\centering\arraybackslash}m{0.25in}}
\multirow{2}{*}{$T_{sal}$} 
& \multicolumn{7}{c}{Level} \\ & 1 & 2 & 3 & 4 & 5 & 6 & 7\\ \hline
20\% & \textbf{100\%} & \textbf{98.67\%} & \underline{92.55\%} & 83.24\% & 74.44\% & 39.88\% & 0.1\% \\
50\% & \textbf{100\%} & 97.95\% & 90.9\% & \textbf{84.2\%} & \textbf{78.54\%} & \underline{66.3\%} & \underline{5.3\%} \\
80\% & \textbf{100\%} & \underline{98.64\%} & \textbf{92.60\%} & \underline{84.17\%} & \underline{76.99\%} & \textbf{69.18\%} & \textbf{22.13\%} \\
\end{tabular}
\vspace{-0.1in}
\caption{
\textbf{Graph coverage vs. control precision (captioning).}
Higher values of $T_{sal}$ allow more diverse descriptions in the self-generated dataset, enhancing the control over long descriptions.
}
\label{table:t_sal}
\end{table}

\noindent
{\bf Data selection parameters.}
Our approach has three parameters:
(1) $c$ that determines $T_i$, i.e. the filtration rate between consecutive iterations; 
(2) $\eta$ that sets the duration (number of steps) of training the model at every iteration;
%(number of training iterations).
% As a consequence, is determines 
and (3) $s$ from Eq.~\ref{eq:step}, that determines the smoothness of the step function that sets the sampling weight of a data point.

The smaller {\em $c$} is, the more gradual  filtration  is.
We observed a similar performance gain when dropping $1$-$5\%$ of the data.
Thus, we set  $c=2\%$.
%

% figure
\begin{figure}[tb]
%\vspace{-0.1in}
\centering
\begin{tabular}{c}
     %\hspace{-0.4cm}
     \includegraphics[width=0.72\linewidth]{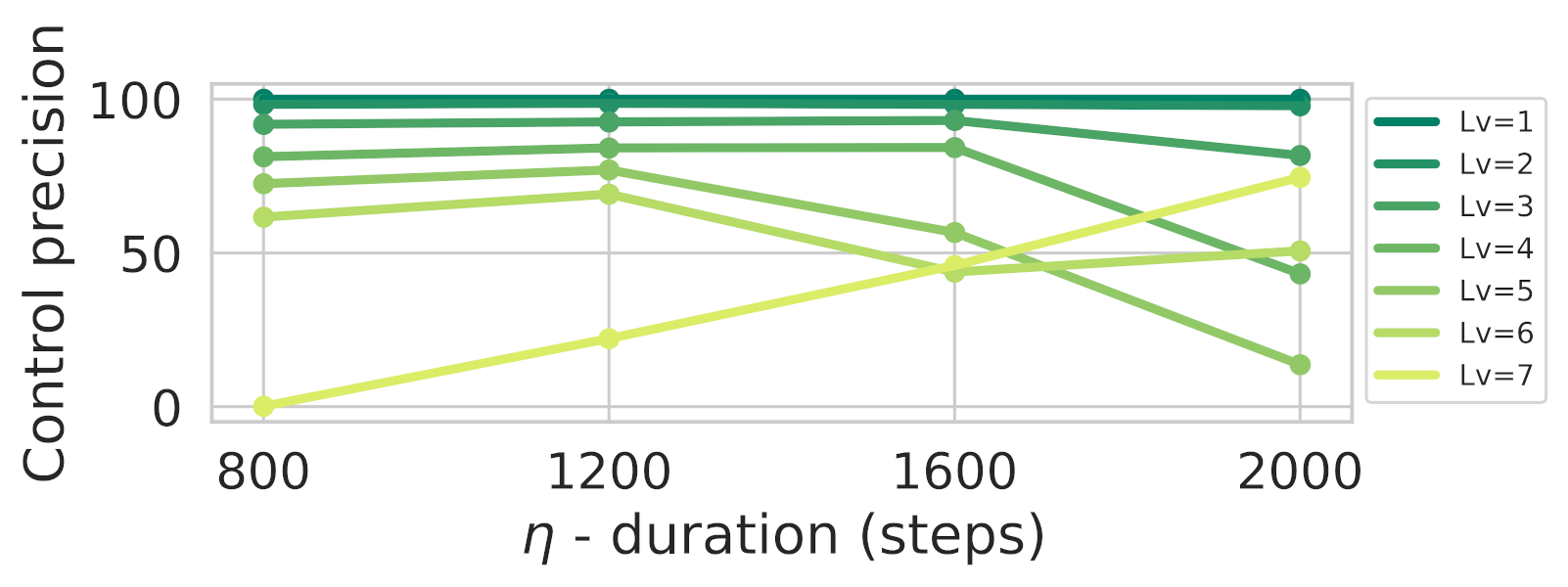} \\
     %\vspace{-0.06in}
     (a) Control precision (\%) vs. $\eta$ \\[0.05in]
     %\hspace{-0.4cm}
     \includegraphics[width=0.72\linewidth]{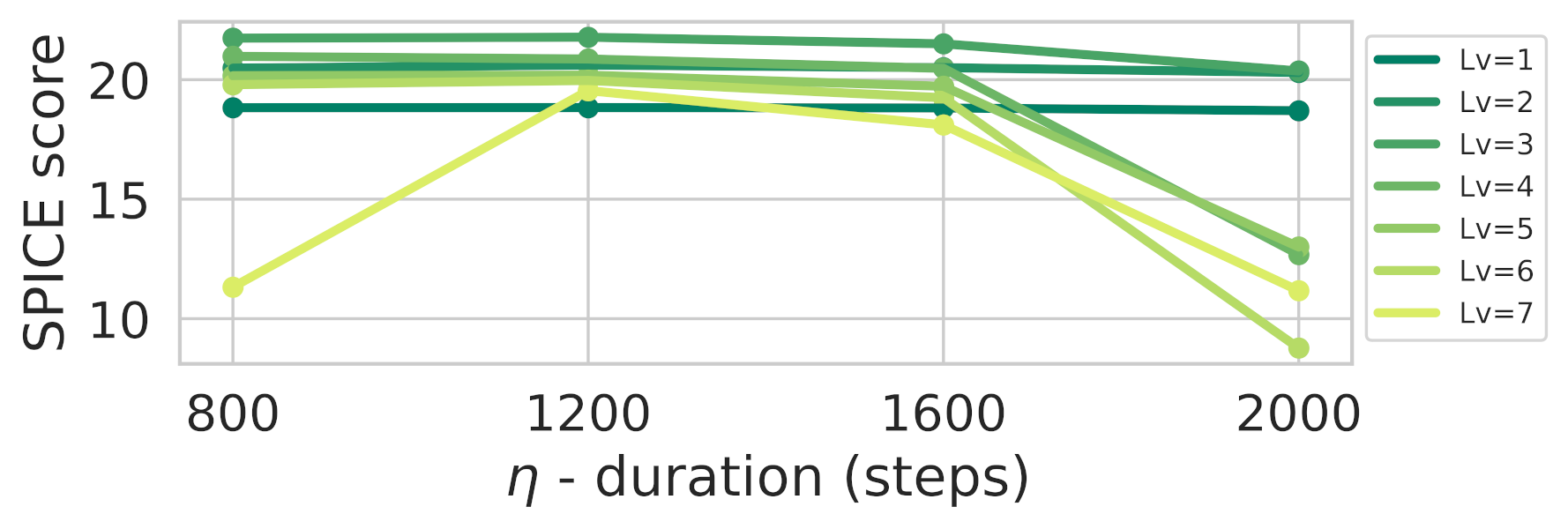} \\
     %\vspace{-0.06in}
      (b) SPICE score vs. $\eta$ 
\end{tabular}
\vspace{-0.08in}
\caption{\textbf{Ablation study of the $\eta$ parameter}.
(a) Control precision and (b) SPICE score per length level (lighter color = longer captions) as a function of the duration $\eta$ at %the step smoothness value 
$s=1$.
%The results are the mean of three runs.
}
\label{fig:ablation}
\vspace{-0.1in}
\end{figure}

% $\eta$ that determines the duration of training the model at every iteration, and thus
% %(number of training iterations).
% % As a consequence, is determines 
% the occurrence of the data points, mostly of the self-generated ones, which are sampled at every stage.
If  {\em $\eta$} is large, sampled data points at a certain iteration are repeated many times, and therefore have more influence on the model.
If {\em $\eta$} is small, each iteration is shorter and the self-generated data is sampled more often, potentially exposing the model to more diverse data, since at every iteration new data is selected. 
%
% The sampled data diversity at every iteration is controlled by {\em $s$}.
{\em $s$} determines the sampled data diversity at every iteration.
% which is the smoothness of the step function used to weight the data points.
%As shown in Fig.~\ref{fig:smoothstep}, 
A small ~$s$ restricts the randomness of data sampling.
% When $s \rightarrow 0$, the data selection process only filters our data points from one stage to the other.
Large values of $s$ give more probability to data points with low-quality to appear at later stages, leading to more diverse data points between consecutive iterations.

We observed a connection between the values of $\eta$ and $s$. 
Increasing $\eta$ usually requires decreasing $s$.
This is logical, since substantial changes in data (high $s$) during consecutive long iterations (high $\eta$) can resemble training the model repeatedly on different domains, potentially causing deviation from the desired domain represented by trusted data.
The effect of high values of $s$ is enhanced when replacing the smooth step function with a constant function (assigning the same probability to all data points at every iteration). 
%This leads to a decline in the output quality for levels $4$-$6$, where data is limited, by as much as $40\%$.
This results in up to a $40\%$ decline in output quality for levels $4$-$6$, where data is limited.

To study the effect of $\eta$ \& $s$, we experiment with $\eta=\{ 800, 1200, 1600, 2000\}$ and $s=\{0.1, 0.5, 1, 1.4\}$.
Levels $1$-$3$, which are $99.8\%$ of the trusted captioning dataset, are hardly affected by these parameters.
Figure~\ref{fig:ablation} shows the influence of $\eta$ on our results, for  $s=1$.
As shown, short captions are barely affected by changing $\eta$.
For the long captions that appear in the dataset (Levels $4$-$6$), there is a sweet-spot at $\eta=1200$ in terms of control and quality.
For level $7$, which does not exist in the trusted dataset, we observe % catastrophic forgetting (Fig.~\ref{fig:ablation}(b)), which is expressed by 
a low control precision for small $\eta$.
Increasing $\eta$ solves this problem.
%, but in turn, degrades the quality (SPICE).
We set $s=1$ and $\eta=1200$ for the best overall performance ($\eta=200$ for paragraph generation).

\noindent
\textbf{Limitations.}
Levels that are completely out-of-distribution ($7$ or up) do not always produce sought-after caption quality. 
% Generating highly detailed descriptions for very simple images is still a limitation of our work.

%%%%%%%%%%%%%%%%%%%%%%%%%%%
%% section: Conclusions
%%%%%%%%%%%%%%%%%%%%%%%%%%%
\vspace{-0.1in}
\section{Conclusions}

We propose a novel, general and unified method to address the shortage of certain length descriptions in common image captioning \& paragraph generation datasets.
Our approach consists of two complementary ideas.
First, we show how to enrich the existing dataset with self-generated varying-length descriptions, using scene graphs and saliency maps.
Second, we introduce a training procedure that gets both a trusted (original) dataset and a self-generated one.
It gradually trains the model so as to learn from the (varying-length) low-quality data, while not harming the information learned from the clean trusted data.

We show that the length-control abilities  vastly improve.
Our method achieves $10\%$ (\slash $14\%$) improvement on average in caption (\slash paragraph) control precision.
This is done while preserving the quality of the output  across all length levels.

\vspace{0.05in}
\noindent
{\bf Acknowledgements.}
We gratefully acknowledge the support of the Israel Science Foundation 2329/22.

\newpage
{\small
\bibliographystyle{ieee_fullname}
\bibliography{egbib}

\begin{thebibliography}{10}\itemsep=-1pt

\bibitem{alikhani2020clue}
Malihe Alikhani, Piyush Sharma, Shengjie Li, Radu Soricut, and Matthew Stone.
\newblock Clue: Cross-modal coherence modeling for caption generation.
\newblock {\em arXiv preprint arXiv:2005.00908}, 2020.

\bibitem{anderson2016spice}
Peter Anderson, Basura Fernando, Mark Johnson, and Stephen Gould.
\newblock Spice: Semantic propositional image caption evaluation.
\newblock In {\em European conference on computer vision}, pages 382--398. Springer, 2016.

\bibitem{anderson2018bottom}
Peter Anderson, Xiaodong He, Chris Buehler, Damien Teney, Mark Johnson, Stephen Gould, and Lei Zhang.
\newblock Bottom-up and top-down attention for image captioning and visual question answering.
\newblock In {\em Proceedings of the IEEE conference on computer vision and pattern recognition}, pages 6077--6086, 2018.

\bibitem{axelrod2017cynical}
Amittai Axelrod.
\newblock Cynical selection of language model training data.
\newblock {\em arXiv preprint arXiv:1709.02279}, 2017.

\bibitem{banerjee2005meteor}
Satanjeev Banerjee and Alon Lavie.
\newblock Meteor: An automatic metric for mt evaluation with improved correlation with human judgments.
\newblock In {\em Proceedings of the acl workshop on intrinsic and extrinsic evaluation measures for machine translation and/or summarization}, pages 65--72, 2005.

\bibitem{chatterjee2018diverse}
Moitreya Chatterjee and Alexander~G Schwing.
\newblock Diverse and coherent paragraph generation from images.
\newblock In {\em Proceedings of the European conference on computer vision (ECCV)}, pages 729--744, 2018.

\bibitem{chen2021human}
Long Chen, Zhihong Jiang, Jun Xiao, and Wei Liu.
\newblock Human-like controllable image captioning with verb-specific semantic roles.
\newblock In {\em Proceedings of the IEEE/CVF Conference on Computer Vision and Pattern Recognition}, pages 16846--16856, 2021.

\bibitem{chen2020say}
Shizhe Chen, Qin Jin, Peng Wang, and Qi Wu.
\newblock Say as you wish: Fine-grained control of image caption generation with abstract scene graphs.
\newblock In {\em Proceedings of the IEEE/CVF Conference on Computer Vision and Pattern Recognition}, pages 9962--9971, 2020.

\bibitem{chen2018factual}
Tianlang Chen, Zhongping Zhang, Quanzeng You, Chen Fang, Zhaowen Wang, Hailin Jin, and Jiebo Luo.
\newblock ``factual''or``emotional'': Stylized image captioning with adaptive learning and attention.
\newblock In {\em Proceedings of the European Conference on Computer Vision (ECCV)}, pages 519--535, 2018.

\bibitem{chen2015microsoft}
Xinlei {Chen}, Hao {Fang}, Tsung-Yi {Lin}, Ramakrishna {Vedantam}, Saurabh {Gupta}, Piotr {Dollar}, and C.~Lawrence {Zitnick}.
\newblock Microsoft coco captions: Data collection and evaluation server.
\newblock {\em arXiv preprint arXiv:1504.00325}, 2015.

\bibitem{chen2019learning}
Yuhua Chen, Wen Li, Xiaoran Chen, and Luc~Van Gool.
\newblock Learning semantic segmentation from synthetic data: A geometrically guided input-output adaptation approach.
\newblock In {\em Proceedings of the IEEE/CVF Conference on Computer Vision and Pattern Recognition}, pages 1841--1850, 2019.

\bibitem{cornia2019show}
Marcella Cornia, Lorenzo Baraldi, and Rita Cucchiara.
\newblock Show, control and tell: A framework for generating controllable and grounded captions.
\newblock In {\em Proceedings of the IEEE/CVF Conference on Computer Vision and Pattern Recognition}, pages 8307--8316, 2019.

\bibitem{cornia2020meshed}
Marcella Cornia, Matteo Stefanini, Lorenzo Baraldi, and Rita Cucchiara.
\newblock Meshed-memory transformer for image captioning.
\newblock In {\em Proceedings of the IEEE/CVF Conference on Computer Vision and Pattern Recognition}, pages 10578--10587, 2020.

\bibitem{deng2020length}
Chaorui Deng, Ning Ding, Mingkui Tan, and Qi Wu.
\newblock Length-controllable image captioning.
\newblock In {\em Computer Vision--ECCV 2020: 16th European Conference, Glasgow, UK, August 23--28, 2020, Proceedings, Part XIII 16}, pages 712--729. Springer, 2020.

\bibitem{deshpande2019fast}
Aditya Deshpande, Jyoti Aneja, Liwei Wang, Alexander~G Schwing, and David Forsyth.
\newblock Fast, diverse and accurate image captioning guided by part-of-speech.
\newblock In {\em Proceedings of the IEEE/CVF Conference on Computer Vision and Pattern Recognition}, pages 10695--10704, 2019.

\bibitem{devlin-etal-2019-bert}
Jacob Devlin, Ming-Wei Chang, Kenton Lee, and Kristina Toutanova.
\newblock {BERT}: Pre-training of deep bidirectional transformers for language understanding.
\newblock In {\em Proceedings of the 2019 Conference of the North {A}merican Chapter of the Association for Computational Linguistics: Human Language Technologies, Volume 1 (Long and Short Papers)}, pages 4171--4186. Association for Computational Linguistics, 2019.

\bibitem{donahue2015long}
Jeffrey Donahue, Lisa Anne~Hendricks, Sergio Guadarrama, Marcus Rohrbach, Subhashini Venugopalan, Kate Saenko, and Trevor Darrell.
\newblock Long-term recurrent convolutional networks for visual recognition and description.
\newblock In {\em Proceedings of the IEEE conference on computer vision and pattern recognition}, pages 2625--2634, 2015.

\bibitem{farhadi2010every}
Ali Farhadi, Mohsen Hejrati, Mohammad~Amin Sadeghi, Peter Young, Cyrus Rashtchian, Julia Hockenmaier, and David Forsyth.
\newblock Every picture tells a story: Generating sentences from images.
\newblock In {\em European conference on computer vision}, pages 15--29. Springer, 2010.

\bibitem{gao2018image}
Lizhao Gao, Bo Wang, and Wenmin Wang.
\newblock Image captioning with scene-graph based semantic concepts.
\newblock In {\em Proceedings of the 2018 10th International Conference on Machine Learning and Computing}, pages 225--229, 2018.

\bibitem{ghosh2019generating}
Shalini Ghosh, Giedrius Burachas, Arijit Ray, and Avi Ziskind.
\newblock Generating natural language explanations for visual question answering using scene graphs and visual attention.
\newblock {\em arXiv preprint arXiv:1902.05715}, 2019.

\bibitem{gu2019unpaired}
Jiuxiang Gu, Shafiq Joty, Jianfei Cai, Handong Zhao, Xu Yang, and Gang Wang.
\newblock Unpaired image captioning via scene graph alignments.
\newblock In {\em Proceedings of the IEEE/CVF International Conference on Computer Vision}, pages 10323--10332, 2019.

\bibitem{han2021image}
Xiaotian Han, Jianwei Yang, Houdong Hu, Lei Zhang, Jianfeng Gao, and Pengchuan Zhang.
\newblock Image scene graph generation (sgg) benchmark, 2021.

\bibitem{hendrycks2018using}
Dan Hendrycks, Mantas Mazeika, Duncan Wilson, and Kevin Gimpel.
\newblock Using trusted data to train deep networks on labels corrupted by severe noise.
\newblock {\em Advances in neural information processing systems}, 31, 2018.

\bibitem{Herdade2019transforming}
Simao Herdade, Armin Kappeler, Kofi Boakye, and Joao Soares.
\newblock Image captioning: Transforming objects into words.
\newblock In H. Wallach, H. Larochelle, A. Beygelzimer, F. d\textquotesingle Alch\'{e}-Buc, E. Fox, and R. Garnett, editors, {\em Advances in Neural Information Processing Systems}, volume~32. Curran Associates, Inc., 2019.

\bibitem{hodosh2013framing}
Micah Hodosh, Peter Young, and Julia Hockenmaier.
\newblock Framing image description as a ranking task: Data, models and evaluation metrics.
\newblock {\em Journal of Artificial Intelligence Research}, 47:853--899, 2013.

\bibitem{huggingface2021pegasus}
HuggingFace.
\newblock Pegasus paraphraser.
\newblock \url{https://huggingface.co/tuner007/pegasus_paraphrase}, 2020.

\bibitem{JIA20EML}
Sen Jia and Neil~D.B. Bruce.
\newblock Eml-net: An expandable multi-layer network for saliency prediction.
\newblock {\em Image and Vision Computing}, 95:103887, 2020.

\bibitem{johnson2018image}
Justin Johnson, Agrim Gupta, and Li Fei-Fei.
\newblock Image generation from scene graphs.
\newblock In {\em Proceedings of the IEEE conference on computer vision and pattern recognition}, pages 1219--1228, 2018.

\bibitem{johnson2015image}
Justin Johnson, Ranjay Krishna, Michael Stark, Li-Jia Li, David Shamma, Michael Bernstein, and Li Fei-Fei.
\newblock Image retrieval using scene graphs.
\newblock In {\em Proceedings of the IEEE conference on computer vision and pattern recognition}, pages 3668--3678, 2015.

\bibitem{karpathy2015deep}
Andrej Karpathy and Li Fei-Fei.
\newblock Deep visual-semantic alignments for generating image descriptions.
\newblock In {\em Proceedings of the IEEE conference on computer vision and pattern recognition}, pages 3128--3137, 2015.

\bibitem{kastner2021imageability}
Marc~A Kastner, Kazuki Umemura, Ichiro Ide, Yasutomo Kawanishi, Takatsugu Hirayama, Keisuke Doman, Daisuke Deguchi, Hiroshi Murase, and Shin’ichi Satoh.
\newblock Imageability-and length-controllable image captioning.
\newblock {\em IEEE Access}, 2021.

\bibitem{kim2019dense}
Dong-Jin Kim, Jinsoo Choi, Tae-Hyun Oh, and In~So Kweon.
\newblock Dense relational captioning: Triple-stream networks for relationship-based captioning.
\newblock In {\em Proceedings of the IEEE/CVF Conference on Computer Vision and Pattern Recognition}, pages 6271--6280, 2019.

\bibitem{krause2016paragraphs}
Jonathan Krause, Justin Johnson, Ranjay Krishna, and Li Fei-Fei.
\newblock A hierarchical approach for generating descriptive image paragraphs.
\newblock In {\em Computer Vision and Pattern Recognition (CVPR)}, 2017.

\bibitem{krishna2017visual}
Ranjay Krishna, Yuke Zhu, Oliver Groth, Justin Johnson, Kenji Hata, Joshua Kravitz, Stephanie Chen, Yannis Kalantidis, Li-Jia Li, David~A Shamma, et~al.
\newblock Visual genome: Connecting language and vision using crowdsourced dense image annotations.
\newblock {\em International journal of computer vision}, 123(1):32--73, 2017.

\bibitem{li2017learning}
Yuncheng Li, Jianchao Yang, Yale Song, Liangliang Cao, Jiebo Luo, and Li-Jia Li.
\newblock Learning from noisy labels with distillation.
\newblock In {\em Proceedings of the IEEE International Conference on Computer Vision}, pages 1910--1918, 2017.

\bibitem{liang2017recurrent}
Xiaodan Liang, Zhiting Hu, Hao Zhang, Chuang Gan, and Eric~P Xing.
\newblock Recurrent topic-transition gan for visual paragraph generation.
\newblock In {\em Proceedings of the IEEE international conference on computer vision}, pages 3362--3371, 2017.

\bibitem{lin2004rouge}
Chin-Yew Lin.
\newblock Rouge: A package for automatic evaluation of summaries.
\newblock In {\em Text summarization branches out}, pages 74--81, 2004.

\bibitem{liu2020peer}
Yang Liu and Hongyi Guo.
\newblock Peer loss functions: Learning from noisy labels without knowing noise rates.
\newblock In {\em International Conference on Machine Learning}, pages 6226--6236. PMLR, 2020.

\bibitem{lu2018neural}
Jiasen Lu, Jianwei Yang, Dhruv Batra, and Devi Parikh.
\newblock Neural baby talk.
\newblock In {\em Proceedings of the IEEE conference on computer vision and pattern recognition}, pages 7219--7228, 2018.

\bibitem{luo2019curiosity}
Yadan Luo, Zi Huang, Zheng Zhang, Ziwei Wang, Jingjing Li, and Yang Yang.
\newblock Curiosity-driven reinforcement learning for diverse visual paragraph generation.
\newblock In {\em Proceedings of the 27th ACM International Conference on Multimedia}, pages 2341--2350, 2019.

\bibitem{mao2018show}
Yuzhao Mao, Chang Zhou, Xiaojie Wang, and Ruifan Li.
\newblock Show and tell more: Topic-oriented multi-sentence image captioning.
\newblock In {\em IJCAI}, pages 4258--4264, 2018.

\bibitem{mathews2018semstyle}
Alexander Mathews, Lexing Xie, and Xuming He.
\newblock Semstyle: Learning to generate stylised image captions using unaligned text.
\newblock In {\em Proceedings of the IEEE Conference on Computer Vision and Pattern Recognition}, pages 8591--8600, 2018.

\bibitem{mittal2019interactive}
Gaurav Mittal, Shubham Agrawal, Anuva Agarwal, Sushant Mehta, and Tanya Marwah.
\newblock Interactive image generation using scene graphs.
\newblock {\em arXiv preprint arXiv:1905.03743}, 2019.

\bibitem{moore2010intelligent}
Robert~C Moore and William Lewis.
\newblock Intelligent selection of language model training data.
\newblock In {\em Proceedings of the ACL 2010 conference short papers}, pages 220--224, 2010.

\bibitem{mu2020learning}
Jiteng Mu, Weichao Qiu, Gregory~D Hager, and Alan~L Yuille.
\newblock Learning from synthetic animals.
\newblock In {\em Proceedings of the IEEE/CVF Conference on Computer Vision and Pattern Recognition}, pages 12386--12395, 2020.

\bibitem{nguyen2021defense}
Kien Nguyen, Subarna Tripathi, Bang Du, Tanaya Guha, and Truong~Q Nguyen.
\newblock In defense of scene graphs for image captioning.
\newblock In {\em Proceedings of the IEEE/CVF International Conference on Computer Vision}, pages 1407--1416, 2021.

\bibitem{papineni2002bleu}
Kishore Papineni, Salim Roukos, Todd Ward, and Wei-Jing Zhu.
\newblock Bleu: a method for automatic evaluation of machine translation.
\newblock In {\em Proceedings of the 40th annual meeting of the Association for Computational Linguistics}, pages 311--318, 2002.

\bibitem{plummer2015flickr30k}
Bryan~A Plummer, Liwei Wang, Chris~M Cervantes, Juan~C Caicedo, Julia Hockenmaier, and Svetlana Lazebnik.
\newblock Flickr30k entities: Collecting region-to-phrase correspondences for richer image-to-sentence models.
\newblock In {\em Proceedings of the IEEE international conference on computer vision}, pages 2641--2649, 2015.

\bibitem{qi2017online}
Mengshi Qi, Yunhong Wang, and Annan Li.
\newblock Online cross-modal scene retrieval by binary representation and semantic graph.
\newblock In {\em Proceedings of the 25th ACM international conference on Multimedia}, pages 744--752, 2017.

\bibitem{qin2019look}
Yu Qin, Jiajun Du, Yonghua Zhang, and Hongtao Lu.
\newblock Look back and predict forward in image captioning.
\newblock In {\em Proceedings of the IEEE/CVF Conference on Computer Vision and Pattern Recognition}, pages 8367--8375, 2019.

\bibitem{sankaranarayanan2018learning}
Swami Sankaranarayanan, Yogesh Balaji, Arpit Jain, Ser~Nam Lim, and Rama Chellappa.
\newblock Learning from synthetic data: Addressing domain shift for semantic segmentation.
\newblock In {\em Proceedings of the IEEE Conference on Computer Vision and Pattern Recognition}, pages 3752--3761, 2018.

\bibitem{shi2019explainable}
Jiaxin Shi, Hanwang Zhang, and Juanzi Li.
\newblock Explainable and explicit visual reasoning over scene graphs.
\newblock In {\em Proceedings of the IEEE/CVF Conference on Computer Vision and Pattern Recognition}, pages 8376--8384, 2019.

\bibitem{tang2020unbiased}
Kaihua Tang, Yulei Niu, Jianqiang Huang, Jiaxin Shi, and Hanwang Zhang.
\newblock Unbiased scene graph generation from biased training.
\newblock In {\em Proceedings of the IEEE/CVF Conference on Computer Vision and Pattern Recognition}, pages 3716--3725, 2020.

\bibitem{tripathi2019compact}
Subarna Tripathi, Sharath Nittur~Sridhar, Sairam Sundaresan, and Hanlin Tang.
\newblock Compact scene graphs for layout composition and patch retrieval.
\newblock In {\em Proceedings of the IEEE/CVF Conference on Computer Vision and Pattern Recognition Workshops}, pages 0--0, 2019.

\bibitem{vaswani2017attention}
Ashish Vaswani, Noam Shazeer, Niki Parmar, Jakob Uszkoreit, Llion Jones, Aidan~N Gomez, {\L}ukasz Kaiser, and Illia Polosukhin.
\newblock Attention is all you need.
\newblock In {\em Advances in neural information processing systems}, pages 5998--6008, 2017.

\bibitem{vedantam2015cider}
Ramakrishna Vedantam, C Lawrence~Zitnick, and Devi Parikh.
\newblock Cider: Consensus-based image description evaluation.
\newblock In {\em Proceedings of the IEEE conference on computer vision and pattern recognition}, pages 4566--4575, 2015.

\bibitem{veit2017learning}
Andreas Veit, Neil Alldrin, Gal Chechik, Ivan Krasin, Abhinav Gupta, and Serge Belongie.
\newblock Learning from noisy large-scale datasets with minimal supervision.
\newblock In {\em Proceedings of the IEEE conference on computer vision and pattern recognition}, pages 839--847, 2017.

\bibitem{vinyals2015show}
Oriol Vinyals, Alexander Toshev, Samy Bengio, and Dumitru Erhan.
\newblock Show and tell: A neural image caption generator.
\newblock In {\em Proceedings of the IEEE conference on computer vision and pattern recognition}, pages 3156--3164, 2015.

\bibitem{wang2019convolutional}
Jing Wang, Yingwei Pan, Ting Yao, Jinhui Tang, and Tao Mei.
\newblock Convolutional auto-encoding of sentence topics for image paragraph generation.
\newblock {\em arXiv preprint arXiv:1908.00249}, 2019.

\bibitem{wang2020cross}
Sijin Wang, Ruiping Wang, Ziwei Yao, Shiguang Shan, and Xilin Chen.
\newblock Cross-modal scene graph matching for relationship-aware image-text retrieval.
\newblock In {\em Proceedings of the IEEE/CVF Winter Conference on Applications of Computer Vision}, pages 1508--1517, 2020.

\bibitem{wang2018denoising}
Wei Wang, Taro Watanabe, Macduff Hughes, Tetsuji Nakagawa, and Ciprian Chelba.
\newblock Denoising neural machine translation training with trusted data and online data selection.
\newblock In {\em Proceedings of the Third Conference on Machine Translation: Research Papers}, pages 133--143. Association for Computational Linguistics, 2018.

\bibitem{wang2017skeleton}
Yufei Wang, Zhe Lin, Xiaohui Shen, Scott Cohen, and Garrison~W Cottrell.
\newblock Skeleton key: Image captioning by skeleton-attribute decomposition.
\newblock In {\em Proceedings of the IEEE conference on computer vision and pattern recognition}, pages 7272--7281, 2017.

\bibitem{xiao2015learning}
Tong Xiao, Tian Xia, Yi Yang, Chang Huang, and Xiaogang Wang.
\newblock Learning from massive noisy labeled data for image classification.
\newblock In {\em Proceedings of the IEEE conference on computer vision and pattern recognition}, pages 2691--2699, 2015.

\bibitem{xu2017scene}
Danfei Xu, Yuke Zhu, Christopher~B Choy, and Li Fei-Fei.
\newblock Scene graph generation by iterative message passing.
\newblock In {\em Proceedings of the IEEE conference on computer vision and pattern recognition}, pages 5410--5419, 2017.

\bibitem{xu2015show}
Kelvin Xu, Jimmy Ba, Ryan Kiros, Kyunghyun Cho, Aaron Courville, Ruslan Salakhudinov, Rich Zemel, and Yoshua Bengio.
\newblock Show, attend and tell: Neural image caption generation with visual attention.
\newblock In {\em International conference on machine learning}, pages 2048--2057. PMLR, 2015.

\bibitem{yang2020hierarchical}
Xu Yang, Chongyang Gao, Hanwang Zhang, and Jianfei Cai.
\newblock Hierarchical scene graph encoder-decoder for image paragraph captioning.
\newblock In {\em Proceedings of the 28th ACM International Conference on Multimedia}, pages 4181--4189, 2020.

\bibitem{yang2019auto}
Xu Yang, Kaihua Tang, Hanwang Zhang, and Jianfei Cai.
\newblock Auto-encoding scene graphs for image captioning.
\newblock In {\em Proceedings of the IEEE/CVF Conference on Computer Vision and Pattern Recognition}, pages 10685--10694, 2019.

\bibitem{young2014image}
Peter Young, Alice Lai, Micah Hodosh, and Julia Hockenmaier.
\newblock From image descriptions to visual denotations: New similarity metrics for semantic inference over event descriptions.
\newblock {\em Transactions of the Association for Computational Linguistics}, 2:67--78, 2014.

\bibitem{zha2019context}
Zheng-Jun Zha, Daqing Liu, Hanwang Zhang, Yongdong Zhang, and Feng Wu.
\newblock Context-aware visual policy network for fine-grained image captioning.
\newblock {\em IEEE transactions on pattern analysis and machine intelligence}, 2019.

\bibitem{zhang2019graphical}
Ji Zhang, Kevin~J Shih, Ahmed Elgammal, Andrew Tao, and Bryan Catanzaro.
\newblock Graphical contrastive losses for scene graph parsing.
\newblock In {\em Proceedings of the IEEE/CVF Conference on Computer Vision and Pattern Recognition}, pages 11535--11543, 2019.

\bibitem{zhang2021vinvl}
Pengchuan Zhang, Xiujun Li, Xiaowei Hu, Jianwei Yang, Lei Zhang, Lijuan Wang, Yejin Choi, and Jianfeng Gao.
\newblock Vinvl: Revisiting visual representations in vision-language models.
\newblock In {\em Proceedings of the IEEE/CVF Conference on Computer Vision and Pattern Recognition}, pages 5579--5588, 2021.

\bibitem{zhang-etal-2019-curriculum}
Xuan Zhang, Pamela Shapiro, Gaurav Kumar, Paul McNamee, Marine Carpuat, and Kevin Duh.
\newblock Curriculum learning for domain adaptation in neural machine translation.
\newblock In {\em Proceedings of the 2019 Conference of the North {A}merican Chapter of the Association for Computational Linguistics: Human Language Technologies, Volume 1 (Long and Short Papers)}, pages 1903--1915. Association for Computational Linguistics, 2019.

\bibitem{zhang2018generalized}
Zhilu Zhang and Mert~R Sabuncu.
\newblock Generalized cross entropy loss for training deep neural networks with noisy labels.
\newblock In {\em 32nd Conference on Neural Information Processing Systems (NeurIPS)}, 2018.

\bibitem{zhong2020comprehensive}
Yiwu Zhong, Liwei Wang, Jianshu Chen, Dong Yu, and Yin Li.
\newblock Comprehensive image captioning via scene graph decomposition.
\newblock In {\em European Conference on Computer Vision}, pages 211--229. Springer, 2020.

\end{thebibliography}
}

\end{document}